\RequirePackage{fix-cm}
\documentclass[smallextended]{svjour3}       % onecolumn (second format)
\smartqed  % flush right qed marks, e.g. at end of proof
\usepackage{graphicx}
\usepackage{algorithm}
\usepackage{algpseudocode}

\usepackage[]{fontenc}
\usepackage[utf8]{inputenc}
\usepackage{soul}
\usepackage{verbatim}
\usepackage{amsmath}
\usepackage{amssymb}
\usepackage{dsfont}
\usepackage{amsfonts}
\usepackage{mathrsfs}
\usepackage{graphicx}
\usepackage{color}
\usepackage{ulem}
\usepackage{xspace,bbm,epic,eepic}
\usepackage{pgfplots}
\usepackage{setspace}
\usepackage{xspace}

\usepackage{ stmaryrd }
\usepackage{graphicx}
\usepackage{lineno}
\usepackage{color}
\usepackage{url}
\usepackage{graphicx}
\usepackage{ stmaryrd }
\usepackage{algorithm}
\usepackage{algpseudocode}

\usepackage{amsmath,amssymb,color} % define this before the line numbering.
\newcommand{\Rr}{\mathds{R}}
\newcommand{\Cr}{\mathds{C}}

\newcommand{\R}{P}

\newcommand{\Q}{Q}

\newcommand{\XXX}{\mathbf{X}} 
\newcommand{\YYY}{\mathbf{Y}} 
\newcommand{\AAA}{{\mathbf{X}}} 
\newcommand{\BBB}{{\mathbf{Y}}} 
\newcommand{\ZZZ}{\mathbf{Z}}

\newcommand{\eg}{\textit{e.g.}, }
\newcommand{\ie}{\textit{i.e.}, }

\DeclareMathOperator*{\argmin}{arg\,min}

\begin{document}

\title{   Low-Rank Dynamic Mode Decomposition:%\thanks{Grants or other notes
%about the article that should go on the front page should be
%placed here. General acknowledgments should be placed at the end of the article.}
}
\subtitle{An Exact and Tractable Solution}

%\titlerunning{Short form of title}        % if too long for running head

\author{Patrick H\'eas          \and
        C\'edric Herzet  %etc.
}
%\authorrunning{Short form of author list} % if too long for running head

\institute{INRIA Centre Rennes - Bretagne Atlantique \&
		IRMAR - UMR CNRS 6625,\\
		campus universitaire de Beaulieu, 35042 Rennes, France.\\
              \email{patrick.heas@inria.fr}           %  \\
%             \emph{Present address:} of F. Author  %  if needed
          }

\date{Received: date / Accepted: date}
% The correct dates will be entered by the editor

\maketitle

\begin{abstract}
This work studies the linear  approximation of high-dimensional dynamical systems using low-rank dynamic mode decomposition (DMD). Searching this approximation in a data-driven approach is formalized as attempting to solve  a low-rank constrained optimization problem. This problem is non-convex and state-of-the-art algorithms are all sub-optimal. This paper shows that there exists a closed-form solution, which is computed  in polynomial time, and characterizes the  $\ell_2$-norm of the optimal approximation error.  The paper also proposes low-complexity algorithms  building  reduced models from this optimal solution, based on singular value decomposition  or eigen value decomposition.  The  algorithms are evaluated by numerical simulations using synthetic and physical data benchmarks.     
\keywords{Reduced models \and low-rank approximation \and constrained optimization \and dynamical mode decomposition}
% \PACS{PACS code1 \and PACS code2 \and more}
% \subclass{MSC code1 \and MSC code2 \and more}
\end{abstract}

\section{Introduction} \vspace{-0.15cm}

\subsection{Context}\label{sec:context}
The numerical discretization of a  partial differential equation parametrized by its initial condition often leads to a very high dimensional system  {of the form:} \vspace{-0.cm}
\begin{align}\label{eq:model_init} 
 \left\{\begin{aligned}
& x_{t}(\theta)= f_t(x_{t-1}(\theta)) \\%\mbox{\quad$\forall\, s\in\Sc,t\in\Tc$},
&x_1(\theta)={\theta}
\end{aligned}\right. ,
\quad t = 2 ,\ldots, T,\vspace{-0.cm}%\\
\end{align} 
\noindent
 {where}  $x_t(\theta)\in \Rr^n$ is the state variable,   $f_t:\Rr^n \to \Rr^n$, and  $\theta \in \Rr^n$ denotes an initial condition. % and $\Sc$ and $\Tc$ are some spatial and temporal resolution domain.
In some context, \eg for uncertainty quantification purposes, one is interested by  computing  a set of  trajectories corresponding to different initial conditions $\theta \in \Theta \subset \Rr^n$. This may constitute an intractable task due to the  high dimensionality of the space embedding the trajectories. For instance, in the case where $f_t$ is linear, the  complexity   required to compute a trajectory of  model \eqref{eq:model_init}   scales in $\mathcal{O}(Tn^2)$, which is prohibitive for large values of  $n$ or  $T$.

 To deal with these large values, reduced models approximate  the trajectories of the system for a range of regimes determined by a set of initial conditions~\cite{2015arXiv150206797C}. 
 A common assumption is that the trajectories of interest are well approximated in a low-dimensional subspace of $\Rr^n$. In this spirit, many  tractable approximations of model \eqref{eq:model_init}  have been proposed, in particular the well-known  {\it Petrov-Galerkin projection}~\cite{quarteroni2015reduced}. 
However, these methods require   the knowledge of the equations ruling the high-dimensional system. 

Alternatively, there exist   data-driven approaches. In particular, linear inverse modeling~\cite{penland1993prediction}, 
 principal oscillating patterns~\cite{Hasselmann88}, or more recently, dynamic mode decomposition (DMD)~\cite{Chen12,2016Dawson,hemati2017biasing,Jovanovic12,kutz2016dynamic,Schmid10,Tu2014391} propose to  approximate the unknown function $f_t$ by a linear and low-rank operator. This linear framework  has been extended to  quadratic approximations of $f_t$ in~\cite{CuiMarzoukWillcox2014}.  Although  linear approximations are in appearance restrictive,  they have  recently sparked a new surge of interest because they are at the core of the so-called extended DMD or kernel-based  DMD     \cite{budivsic2012applied,li2017extended,williams2015data,williams2014kernel,2017arXiv170806850Y}.  The latter   decompositions   characterize accurately non-linear behaviours under certain conditions~\cite{klus2015numerical}.

%Beyond the reduced modeling motivation, DMD has initially been proposed for the analysis of experimental data \cite{Schmid10}.  Deducing this decomposition from experimental data reveals 
%  spatio-temporal patterns, which may be relevant for understanding the dynamics. However, in many situations, it is not trivial to extract these patterns accurately. This difficulty arizes in particular when the  data to be analysed is corrupted by noize. To deal with such situations, authors in~\cite{Chen12,Jovanovic12} have proposed algorithms to identify the dominant parameter of DMD from noisy data by introducing a low-rank constraint in the DMD parameter estimation process. 

  Reduced models based on  low-rank linear approximations substitute  function $f_t$ by a matrix   $\hat A_k \in \Rr^{n \times n}$ with $r={\textrm{rank}(\hat {A}_k)} \le n$   as
\begin{align}\label{eq:model_koopman_approx} 
 \left\{\begin{aligned}
& \tilde x_{t}(\theta)=  \hat A_k \tilde  x_{t-1}(\theta),\quad t=2,\ldots,T,  \\%\mbox{\quad$\forall\, s\in\Sc,t\in\Tc$},
&\tilde x_1(\theta)={\theta},
\end{aligned}\right. \vspace{-0.cm}%\\
\end{align} 
where  $\{\tilde x_{t}(\theta)\}_{t=1}^T$ denotes an approximation of the trajectory $\{x_t(\theta)\}_{t=1}^T$ of system \eqref{eq:model_init}. 
%The reduced model \eqref{eq:model_koopman_approx} can  
%Obviously, a brute-force evaluation of a trajectory approximation with \eqref{eq:model_koopman_approx} does not induce a low computational cost:  the complexity  scales in $\mathcal{O}(Tn^2)$. However, 
The complexity for the evaluation of a trajectory approximation with \eqref{eq:model_koopman_approx} will be refered to as {\it on-line complexity}. A low on-line complexity is obtained by exploiting  the low rank  of matrix $\hat A_k$.  A  scaling  in  $\mathcal{O}(Tr^2+rn)$ is reached if the reduced model is parametrized by  matrices  $R,\,L \in \Cr^{n \times r}$ and $S \in \Cr^{r \times r}$  such that  trajectories of \eqref{eq:model_koopman_approx} correspond to  the  recursion
\begin{equation}\label{eq:genericROM0}
 \left\{\begin{aligned}
& \tilde x_{t}(\theta) =  R  z_t,  \quad t=2,\ldots,T,\\
& z_t= S  z_{t-1},  \,\,\,\quad t=3,\ldots,T,\\
&z_2=L^\intercal \theta. 
\end{aligned}\right.
\end{equation}
%\remCH{Pourquoi compliquer ainsi et ne pas mettre simplement $x_t = R z_t$ pour $t=\textcolor{red}{2}\ldots T$,  $z_t= S z_{t-1}$ pour $t=\textcolor{red}{3}\ldots T$ et $z_{\textcolor{red}{2}} = L^T \theta$ avec $S = Q^T P$, $R=P$, $L=Q$ ? Cela a l'avantage de mieux coller à la formulation de la recursion \eqref{eq:model_init}-\eqref{eq:model_koopman_approx}}
The equivalence of  systems    \eqref{eq:model_koopman_approx} and \eqref{eq:genericROM0} is obtained for   $T\ge 2$ by setting  
$\hat A_k^{T-1}=R S^{T-2}L^\intercal$. 
%The complexity to compute an approximated trajectory with the reduced model~\eqref{eq:genericROM0} scales at most in  $\mathcal{O}(Tk^2+kn)$.
% Nevertheless, the construction of this reduced model presuppose the knowledge of a factorization of the form \eqref{eq:solFactor}. Of course,  matrices $P=U_{\hat A_k}$ and $Q^\intercal=\Sigma_{\hat A_k} V_{\hat A_k}^\intercal$ satisfy \eqref{eq:solFactor} for  $r$ at most equal to $k$.  However, the computation of the SVD of $\hat A_k \in \Rr^{n \times n}$ may be painful since it can require at most a  complexity of $\mathcal{O}(n^3)$. Hopefully, there often exist other possible choices  satisfying \eqref{eq:solFactor} and the complexity can often  be lowered to  $\mathcal{O}(m^2(m+n))$. This is the case for approximations obtained by the truncated approach (see Section \ref{sec:trunc}) if we set $P=U_\YYY$ and $Q^\intercal = \Sigma_\YYY V_\YYY^\intercal \AAA^\dagger$, obtained by projected DMD (see Section \ref{sec:LowRankProj}) if we set $P=U_\AAA$ and $Q^\intercal = \tilde B \Sigma_{\AAA}^{\dagger}U_{\AAA}^\intercal$  or obtained by sparse DMD  (see Section \ref{sec:sparse}) if we set $P=U_\AAA$ and adapt  $Q^\intercal$ consequently. Unfortunately,  regularized approaches of Section \ref{sec:convex} do not naturally exhibit such a factorization.
In particular, consider a factorization of the form 
\begin{align}\label{eq:solFactor}
\hat A_k=PQ^\intercal \quad \textrm{with}\quad P,Q \in \Rr^{n \times r}. %, \quad P^\intercal P =I_r.
\end{align}
 This factorization is always possible by computing the singular value decomposition (SVD)  ${\hat A_k}=U_{\hat A_k}\Sigma_{\hat A_k} V_{\hat A_k}^\intercal$ and identifying $P=U_{\hat A_k}$ and $Q^\intercal =\Sigma_{\hat A_k} V_{\hat A_k}^\intercal$. 
 %Since  $\R$ has orthonormal columns,  
 Factorization~\eqref{eq:solFactor} implies  that trajectories of \eqref{eq:model_koopman_approx} are  obtained with system~\eqref{eq:genericROM0}  setting $R=\R$, $L=\Q$ and $S=  \Q^\intercal  \R$.
 %, or equivalently with the $r$-dimensional recursion~\eqref{eq:genericROM0} \remCH{Il me semble que l'égalité "$\hat A_k=R SL^\intercal$" n'est pas vérifiée avec les définitions que tu donnes pour $R,L,S$. De plus, je pense que c'est plutot $R=\R$, $L=\Q$ and $S= \Q^\intercal  \R$ qui mène à~\eqref{eq:genericROM0}}. %The on-line complexity to compute this recursion is  $\mathcal{O}(r^2T+rn)$.
 Another factorization of interest relies on the eigenvalue decomposition (EVD) 
\begin{align}\label{eq:factorEVD}
\hat A_k = D \Lambda D^{-1},\quad \textrm{with}\quad D,\Lambda \in \Cr^{n \times n},
\end{align}%The obtention of this alternative formulation is also detailed in the  appendix. 
where   $\Lambda$ is a Jordan-block matrix \cite{golub2013matrix}  of rank  $r.$ 
%$\hat A_k =R S R^{-1}$  $S $ is a Jordan-block matrix \cite{golub2013matrix} and  $r={\textrm{rank}(\hat {A}_k)}.$  
%By setting $R$ and $L$ as the first $r$ columns of $D$ and $(D^{-1})^\intercal$, and $S$ as the first $r \times r$ block of $\Lambda$,n a direct equivalence with recursion~\eqref{eq:genericROM0}. 
Using the ``economy size''  EVD yields a system of the form of \eqref{eq:genericROM0}. Indeed, it is obtained by making  the identification 
  $L=(\xi_1\cdots \xi_r) $ and  $R=(\zeta_1\cdots \zeta_r)$, where $\xi_i\in \Cr^n$ and $\zeta_i\in \Cr^n$ are the $i$-th left and right eigenvectors of $\hat A_k$ (equivalently the $i$-th column of $(D^{-1})^\intercal$ and $D$), and identifying $S$ to the first $r \times r$ block of $\Lambda$ multiplied by $L^\intercal$.%  we obtain $\tilde x_{t}(\theta)=R S_{t-1}L^\intercal \theta$.
  %   we rewrite \eqref{eq:factorEVD} as    $\hat A_k = R SL^\intercal$.  

The on-line complexity to compute this recursion is still  $\mathcal{O}(Tr^2+rn)$.  But  assuming that $\hat {A}_k$ is diagonalizable\footnote{Diagonalizability is guaranteed if all the non-zero eigenvalues are distinct. However, this condition is  only sufficient and the class of diagonalizable matrices is  larger \cite{Horn12}.}, we have $S =\textrm{diag}(\lambda_1,\cdots, \lambda_r)$ and system~\eqref{eq:genericROM0} becomes
%\end{align}
%Setting $L=(R^{-1})^\intercal$, we obtai Assuming that $\hat {A}_k$ is diagonalizable\footnote{Diagonalizability is assured if all the non-zero eigenvalues are distinct. However, this condition is  only sufficient and the class of diagonalizable matrices is  larger \cite{Horn12}.}, we have $S =\textrm{diag}(\lambda_1,\cdots, \lambda_r)$ and recursion~\eqref{eq:genericROM0} becomes
\begin{align}\label{eq:koopman1}
\left\{\begin{aligned}
 \tilde x_{t}(\theta)&= \sum_{i=1}^{r}\zeta_i \nu_{i,t},\\
\nu_{i,t}& =  \lambda_i^{t-1} \xi_i^\intercal  \theta, \quad \textrm{for} \quad i=1,\ldots, \textrm{rank}(\hat {A}_k)
\end{aligned}\right. , \quad t=2,\ldots,T,\vspace{-0.cm}%\\
\end{align} 
%\begin{align}
%\left\{\begin{aligned}
% \tilde x_{t}&= \sum_{i=1}^{k'} \nu_{i,t} \mu_i,\label{eq:koopman1}\\
%\nu_{i,t}& =  \lambda_i^{t-1} \tilde \varphi_i(\theta),
%\end{aligned}\right. \vspace{-0.cm}%\\
%\end{align} 
where  $\lambda_i \in \Cr$  is the $i$-th  (non-zero) eigenvalue of $\hat {A}_k$.  This reduced-model possesses  %, depending on the low-rank linear approximation $\hat A_k$. % is an approximation of problem \eqref{eq:prob1} for $k=m$ or $k<m$. % The elements of the triplets $( \xi_i, \zeta_i, \lambda_i)$ are respectively approximations of  the $i$-th {\it Koopman  eigen-function}, {\it   eigen-mode}  and {\it eigenvalue} .  
%Note that in the case $\Psi^{-1}\in \Rr^{p \times n}$, then  the first equation of \eqref{eq:koopman1} simplifies to 
%$$ \tilde x_{t}= \sum_{i=1}^{k'} \nu_{i,t} \mu_i, \quad\textrm{with}\quad  \mu_i=  \Psi^{-1} \zeta_i \in \Cr^p.$$ 
%The $\mu_i$'s are in this case the so-called   {\it  eigenmodes}.
 %This appendix also details the relation of matrix $\hat A_k$ with a finite approximation of the Koopman operator.
%As detailed in Section \ref{sec:context},   \eqref{eq:koopman1} is 
 a very desirable  on-line complexity of  $\mathcal{O}(rn)$,  
%  versus at best    $\mathcal{O}(r^2T+rn)$ for the SVD-based reduced model presented in the previous section.
%on-line complexity of $\mathcal{O}(rn)$,  
\ie  linear in the ambient dimension $n$, linear in the reduced-model intrinsic dimension $r$ and independent of the trajectory length~$T$. \\

The key of reduced modeling  is to find a ``good'' tradeoff between  the  on-line complexity and the  accuracy of the approximation. As shown previously, the  low on-line computational effort is obtained by a proper factorization of the low-rank matrix $\hat A_k$. Thus,  in an off-line stage, it remains  to {\it i)}~search $\hat A_k$ within the family of low-rank matrices which yields the ``best''  approximation~\eqref{eq:model_koopman_approx}, {\it ii)}~compute the SVD or EVD based factorization  of   $\hat A_k$.   We will refer to  the computational cost associated to these two steps as {\it off-line complexity}.

%If $S \in \Cr^{r \times r}$ is a block diagonal  matrix and in particular a Jordan matrix,
%the on-line complexity to run  reduced-model \eqref{eq:genericROM0} scales  in $\mathcal{O}(s^2T+rn)$  where $s$ denotes the maximum size of the  blocks.  In the case  $S$ is a diagonal matrix,  then  trajectories of \eqref{eq:genericROM0} are computed with the  advantageous on-line complexity of $\mathcal{O}(rn)$,  \ie  linear in the ambient dimension $n$, linear in the reduced-model intrinsic dimension $r$ and independent of the trajectory length~$T$.   

%  
%
% As we will see, 
%%by singular value decomposition (SVD) of 
% %$\hat A_k$, the number of operations  necessary to compute an approximated  trajectory will scale at worst in  $\mathcal{O}(k^2T+kn)$. This complexity can be further lowered  to   $\mathcal{O}(s^2T+kn)$ by EVD of $\hat A_k$, where $s$ denotes the maximum size of the Jordan blocks.  
% under certain conditions %in the case   $\hat A_k$ is diagonalizable
% we can reach the advantageous complexity of $\mathcal{O}(kn)$, \ie  linear in the ambient dimension $n$, linear in the reduced-model intrinsic dimension $k$ and independent of the trajectory length~$T$. 

A standard choice is to select $\hat A_k$ inducing   the best  trajectory approximation in the $\ell_2$-norm sense, for initial conditions in the set  $\Theta \subset \Rr^n$:  matrix  $\hat A_k$ in \eqref{eq:model_koopman_approx} targets  the solution of the following minimization problem for some given $k \le n $:\vspace{-0.2cm}
\begin{align}\label{eq:target}
\argmin_{A:\textrm{rank}(A)\le k}  \int_{\theta \in \Theta} \sum_{t=2}^T \| x_{t}(\theta) - A^{t-1} \theta\|^2_2, %\mbox{\quad$\forall\, s\in\Sc,t\in\Tc$},
\end{align}
where $\|\cdot\|_2$ denotes the $\ell_2$-norm. %A reduced model  of the form of \eqref{eq:genericROM0} based on a low-rank minimizer  of~\eqref{eq:target} is then  deduced from its EVD: $R$ and $L$ are set as the  right and left eigenvectors and $S$ as the matrix of eigenvalues. 
%The computational effort necessary to  solve problem \eqref{eq:model_koopman_approx} and to obtain the parameters $L,R$ and $S$. \\%\footnote{ The   {\it on-line complexity} necessary to run the reduced model for a given initial condition  is to be opposed to the {\it off-line complexity}  necessary  to build the reduced model.}   \\
Since we focus on  {\it data-driven} approaches, we  assume that  we do not know the exact form of $f_t$ in~\eqref{eq:model_init}
 and we only have access to a set of representative trajectories $\{x_t(\theta_i)\}_{t=1}^T$, $i=1,...,N$ so-called \textit{snapshots}, obtained by running the high-dimensional system for $N$ different initial conditions $\{\theta_i\}_{i=1}^N$ in the set $\Theta$.   
Using these snapshots, we consider a discretized version of~\eqref{eq:target}, which corresponds to the constrained optimization problem studied  in \cite{Chen12,Jovanovic12,wynn2013optimal}: matrix $\hat A_k$ now targets  the solution %, which we will refer to as the problem of \textit{low-rank DMD estimation}
\begin{align}\label{eq:prob} 
A_k^\star \in &\argmin_{A:\textrm{rank}(A)\le k}  \sum_{i=1}^{N}  \sum_{t=2}^{T} \| x_{t}(\theta_i) - A   x_{t-1}(\theta_i)\|^2_2, %\mbox{\quad$\forall\, s\in\Sc,t\in\Tc$},
\end{align} 
 where we have substituted   $A^{t-1} \theta_i$ in \eqref{eq:target} by   $A   x_{t-1}(\theta_i)$ and where we have approximated the integral by an empirical average over the snapshots. 

 Problem~\eqref{eq:prob} is non-convex due to the presence of the rank constraint ``$\textrm{rank}(A)\le k$''. As consequence, it has  been considered as intractable in several contributions of the litterature and numerous procedures have been proposed to approximate its solution (see next section). 
 % As detailed in the next section, several procedures have been proposed to approximate the solution of this problem. 
 In this paper, we show that problem \eqref{eq:prob} is in fact tractable and admits a closed-form solution which can be evaluated in polynomial-time. \vspace{-0.15cm}
 %shows that there exists an exact closed-form solution.  
% Note that in the case $k \ge N(T-1)$, the solution of \eqref{eq:prob} is simply the solution of the unconstrained  version of this least-square problem. %A solution is in this case  simply obtained by  singular value decomposition (SVD) \cite{Tu2014391}.
%  In what follows, we will focus on the more involved situation where  $k < N(T-1)$.\\
%  

%. This reduced model~\eqref{eq:genericROM0} is simply called { ``DMD''}  in the case $L,\, R$ and $S$ are related to the EVD  truncated to  $k$ terms of $\hat A_k$, set in this case as  the solution of the unconstrained version of problem \eqref{eq:target}.  

%We remark that the overall complexity   necessary to compute a trajectory with the original model \eqref{eq:model_init} scales  in $\mathcal{O}(Tn^2)$ in the case $f_t$ is a linear and full-rank operator. In contrast,  %assuming the  $\mu_i$'s are given and assuming the computation of the $\tilde \varphi_i(\theta)$'s imply a complexity scaling linearly in $n$,
%%and that a matrix vector multiplication by $\Psi^{-1}$ is of order\footnote{{Note that the complexity of this multiplication is  $\mathcal{O}(p)$ if the the columns of $\Psi$  include the canonical basis.}} $\mathcal{O}(p)$,
% %then  
% the reduced model \eqref{eq:koopman1}  will have a complexity of   $\mathcal{O}(kn)$,  \ie scaling linearly (up to the multiplication factor $k$) with the ambient space dimension $n$ and independent of the  trajectory length $T$. 
% 
\subsection{Problem Statement and Contributions}

This work deals with the off-line construction of reduced models of the form of~\eqref{eq:genericROM0}. It focuses on the following  questions:
\begin{enumerate}
  \item Can we compute a solution of problem \eqref{eq:prob}  in polynomial time?
  \item How to compute efficiently a factorization of this solution, and in particular its EVD?
\end{enumerate}
Let us make some correspondences with the terminology used in the DMD literature~\cite{Chen12,2016Dawson,hemati2017biasing,Jovanovic12,kutz2016dynamic,Schmid10,Tu2014391} in order to reformulate these two questions in the jargon used in this community. 
% so that we can reformulate the problem statement in those terms.  % for the  solution \eqref{eq:prob} and the reduced model \eqref{eq:koopman1}.    
%Let columns of $R$ and $L$ be the dominant  right  and left eigenvectors  of $\hat A_k$  and let $S$ be the diagonal matrix gathering the first $k$ eigenvalues.  Given this choice for matrices $R$, $L$ and $S$, 
The   { ``low-rank DMD''}  of system \eqref{eq:model_init} refers to the EVD of the solution $A_k^\star$ of problem \eqref{eq:prob}, or equivalently to the parameters of reduced model~\eqref{eq:koopman1}  in the case where $\hat A_k=A_k^\star$ is diagonalizable.\footnote{The ``DMD''  of system \eqref{eq:model_init} refers to the EVD of the solution of problem \eqref{eq:prob} without the low-rank constraint.}  
% Using this terminology, the two previous questions are summarized in the following problem statement: can we compute exactly and with a polynomial complexity  the low-rank DMD  of system \eqref{eq:model_init}? \\
Using this terminology, the two above questions can be summarized summarized as follows: can we compute exactly and with a polynomial complexity the low-rank DMD  of system \eqref{eq:model_init}? 
In this paper, we show that the answer to this question is positive and provide some numerical procedures to attain this goal. \\

\textbf{Solver for problem \eqref{eq:prob}.}\,
In the last decade, there has been a surge of interest for  low-rank solutions of linear matrix equations, see \eg~\cite{fazel2002matrix,jain2010guaranteed,lee2009guaranteed,lee2010admira,mishra2013low,recht2010guaranteed}.  This  class of  problems includes \eqref{eq:prob} as an important  particular  case. Problems in this class  are always non-convex due to the rank constraint and  computing their solutions in polynomial time  is often out of reach. Nevertheless,   certain instances of these problems with  very special structures admit closed-form solutions \cite{eckart1936approximation,mesbahi1997rank,parrilo2000cone}. In this work, we show that \eqref{eq:prob} belongs to this class of problems and provide a closed-form solution which can be computed in polynomial time. Prior to this work, many authors have proposed tractable procedures to compute approximations of the solution to problem  \eqref{eq:prob}~ \cite{Chen12,Jovanovic12,li2017extended,Tu2014391,wynn2013optimal,2017arXiv170806850Y} or to related problems~\cite{hemati2017biasing}. We review these contributions in Section~\ref{sec:approxLDMD} and discuss their complexity.\vspace{-0.15cm}\\

\textbf{Factorization of the solution.}
The second problem concerns the computation of the factorization of the form \eqref{eq:solFactor} or \eqref{eq:factorEVD} of the solution $A_k^\star  \in \Rr^{n \times n}$. 
%We will see that in most cases it is straightforward to compute a reduced model in the form of a low-dimensional recursion with a complexity of . However, 
%It is not  clear that this will not imply a prohibitive computational burden for large $n$. Indeed the
A brute-force computation of a factorization of a matrix in $\Rr^{n \times n}$, in particular an EVD, is  prohibitive for  large values of $n$. In this work, we propose  low-complexity algorithms  computing such factoization of $A_k^\star$. This follows the line and extends previous works~\cite{Jovanovic12,Tu2014391,williams2014kernel}, as detailed in Section \ref{sec:etatArt2}.\vspace{-0.15cm}\\

In summary, the contribution of this paper  is twofold. First, we provide a closed-form  solution  to~\eqref{eq:prob}. We also design an algorithm  computing a factorized form of this solution with a linear complexity  in the ambient dimension. Second, we provide an algorithm computing  the EVD  of  this optimal  solution, which does not imply an increase in complexity.   \vspace{-0.15cm}\\

The paper is organized as follows. In Section \ref{sec:stateArt}, we %present the Koopman-based reduced-modeling framework and then 
provide a  review of techniques approximating and factorizing the solution of problem \eqref{eq:prob}. In Section~\ref{sec:contrib}, 
we present the proposed approach. 
Finally, in Section~\ref{sec:numEval},   we study  the performance obtained with the proposed algorithms in synthetic and physical setups   and compare with state-of-the-art.
 \vspace{-0.25cm}
 \section{Notations}\label{sec:notations}\vspace{-0.05cm}
 %In what follows,  we assume that we have at our disposal  $N$ trajectories of $T$ snapshots.  
%\remCH{Introduire la definition de la transpose? Dans la suite, tu fais implicitement l'hypothèse que $m\leq n$; il faudrait le dire a ce stade.}
 All along the paper, we   make extensive use    of the  economy-size SVD  of a matrix  $M\in \Rr^{p \times q }$ with $p\ge q$: $M=U_M\Sigma_M V_M^\intercal $ with $U_M\in \Rr^{p \times q }$, $V_M\in \Rr^{ q \times  q}$ and $\Sigma_M\in \Rr^{q  \times q }$ so that $U_M^\intercal U_M=V_M^\intercal V_M=I_q$ and $\Sigma_M $ is diagonal, where the  upper script~$\cdot^\intercal $  refers to the transpose and $I_q$  denotes the $q$-dimensional identity matrix. 
 %\footnote{This ``economy size''  SVD omits the trailing $p-q$ all-zero rows of $\Sigma_M$  and the corresponding columns of $U_M$.}
 The columns of matrices $U_M$ and $V_M$ are  denoted $U_M=(u_M^1 \cdots u_M^q)$ and $V_M=(v_M^1 \cdots v_M^q)$ while  $\Sigma_M  =\textrm{diag}( \sigma_{M,1},  \cdots,  \sigma_{M,q})$  with $\sigma_{M,i} \ge \sigma_{M,i+1}$ for $i=1,\ldots, q-1$. The  Moore-Penrose pseudo-inverse of matrix $M$ is then  defined as $M^{\dagger}=V_M\Sigma^{\dagger}_M U_M^\intercal $, where 
 $\Sigma^{\dagger}_M=\textrm{diag}( \sigma_{M,1}^{\dagger}, \cdots , \sigma_{M,q}^{\dagger})$ with
 $$ 
 \sigma_{M,i}^{\dagger}=  \left\{\begin{aligned}
&\sigma_{M,i}^{-1}\quad \textrm{if}\quad  \sigma_{M,i} > 0\\%\mbox{\quad$\forall\, s\in\Sc,t\in\Tc$},
&0\quad\quad\,\,\,\textrm{otherwise}
\end{aligned}\right. .\vspace{-0.cm}\\
$$
The orthogonal projector onto the span of the columns (resp. of the rows) of matrix $M$ is denoted by $\mathbb{P}_{M}=M M^\dagger=U_M\Sigma_M\Sigma_M^\dagger U_M^\intercal $ (resp. $\mathbb{P}_{M^\intercal}=M^\dagger M=V_M \Sigma_M^\dagger \Sigma_M V_{M}^\intercal$) \cite{golub2013matrix}. 

We also introduce additional notations to derive a matrix formulation of the low-rank estimation problem \eqref{eq:prob}. We gather  consecutive elements of the $i$-th snapshot trajectory between time $t_1$ and $t_2$   in matrix $X_{t_1:t_2}^{(i)}~= ~(x_{t_1}(\theta_i) \cdots x_{t_2}(\theta_i))$ and form large matrices $ \XXX, \YYY \in \Rr^{n \times  m} $ with $m=N(T-1)$  as 
%let  $ C =  \BBB\AAA^\intercal  \quad \textrm{and}\quad  \bar C=  (\AAA\AAA^\intercal )^{\dagger}\AAA\BBB^\intercal ,$  with 
%  \begin{align*}
 $$\XXX = (X^{(1)}_{1:T-1} \cdots  X^{(N)}_{1:T-1})  \quad \textrm{and} \quad 
 \YYY= ( X^{(1)}_{2:T} \cdots X^{(N)}_{2:T}).  $$
% \end{align*}
 In order to be consistent with the SVD definition and to keep the presentation as simple as possible, this work  assumes that $m\leq n$. However, all the result presented in this work can be extended without any difficulty to the case where  $m> n$ by using an alternative definition of the SVD.\vspace{-0.25cm}% {and that $\textrm{rank}(\AAA)\neq 0$ and $\textrm{rank}(\BBB)\neq 0$. }  
		%For simplification issues,  t
		
		%This assumption is in fact not restrictive since it is always possible to remove linear dependence between the columns of  matrices $\AAA$ and $\BBB$   in a pre-processing step  by a proper orthogonalization procedure.
		%We will denote the residual by  $ R= \BBB -A \AAA \in \Rr^{n \times m}$.

%\remCH{Dire qu'on va presenter deux methodes de la litterature qui permettent de calculer une approximation de la solution de \eqref{eq:prob1}.}

\section{State-Of-The-Art Approximations}\label{sec:stateArt}

We begin by  presenting state-of-the-art methods solving  approximatively the low-rank minimization problem \eqref{eq:prob}.  In a second part, we  make an overview of state-of-the-art  algorithms  computing   factorizations of these approximated solutions of the form of  \eqref{eq:solFactor} or \eqref{eq:factorEVD}. \vspace{-0.25cm}

 \subsection{Tractable Approximations to Problem~\eqref{eq:prob}}\label{sec:approxLDMD}
Using   the notations introduced in Section~\ref{sec:notations}, problem~\eqref{eq:prob} can be rewritten as
		\begin{align}\label{eq:prob1} 
		A_k^\star \in &\argmin_{A:\mathrm{rank}(A)\le k} \|\BBB -A \AAA \|^2_F,
		\end{align} 
where $\|\cdot\|_F$ refers to the Frobenius norm. A detailed review of the following state-of-the-art approximations of $A_k^\star$ is provided in our technical note~\cite{HeasHerzet17}. \\

 \textbf{Truncation of the unconstrained solution.}  \label{sec:trunc} A first approximation consists in removing the low-rank constraint in problem~\eqref{eq:prob1}. %, \ie  $k \ge m$,\remCH{Dire plutot, "quand on enleve la contrainte de rang faible". C'est plus explicite que $k\geq m$}
As pointed out by {\it Tu et al.} in \cite{Tu2014391},  the problem then  boils down to a  least-squares  problem 
  		\begin{align}\label{eq:prob1_unconst} 
		\argmin_{A} \|\BBB -A \AAA \|^2_F,
		\end{align} 
		 admitting the closed-form solution $\BBB\AAA^{\dagger}$.
Matrix $\BBB\AAA^{\dagger}$ also solves  the constrained  problem \eqref{eq:prob1} in the case where  $k\ge m$ and in particular  for $k=m$, \ie \vspace{-0.15cm}
\begin{align}\label{eq:exactDMD}
A^\star_m=\BBB\AAA^{\dagger}.
\end{align}
% and in the case where the cost function at this point vanishes (for example in the case where $\AAA$ is full rank).
This solution relies on the  SVD of $\XXX$: $A^\star_m=\BBB V_{\AAA}\Sigma_{\AAA}^{\dagger}U_{\AAA}^\intercal,$ which is computed with a complexity of $\mathcal{O}(m^2(m+n))$~\cite{golub2013matrix}.  An approximation of the solution of~\eqref{eq:prob1} satisfying the low-rank constraint $\textrm{rank}(A)\le k$ with  $k < m$ is then  obtained by a truncation of the SVD or the EVD of $A^\star_m$ using $k$ terms. \\

  \textbf{Approximation by low-rank projected DMD.}\label{sec:LowRankProj}  The {\it ``projected  DMD''} proposed  by {\it Schmid}~\cite{Schmid10} is an approximation of $A^\star_m$, which assumes that the columns of $A^\star_m\AAA$ are in the span of $\AAA $.  This assumption is used by {\it Jovanovic et al.}~\cite{Jovanovic12}   to approximate $A^\star_k$ for $k<m$. Their approach yields the so-called {\it ``low-rank projected  DMD''} approximation of~\eqref{eq:prob1} which takes the following form 
 \begin{align}\label{eq:projDMD}
A_k^\star\approx U_{\AAA}\tilde Y_k \Sigma_{\AAA}^{\dagger}U_{\AAA}^\intercal,
 \end{align}
where $\tilde Y_k$ denotes the SVD representation of matrix $\tilde Y=U_{\AAA}^\intercal \BBB V_{\AAA}$  truncated to $k$ terms.
Similar  low-dimensional parametrizations of the optimal solution $A_k^\star$ are used to compute the so-called {\it ``optimized DMD''} in \cite{Chen12} or  {\it ``optimal mode decomposition''} in \cite{wynn2013optimal}.   
The computation of low-rank projected  DMD relies on the SVD of   $\AAA\in \Rr^{n \times m}$ and $\tilde Y\in \Rr^{m \times m}$ and thus involves a complexity of  $\mathcal{O}(m^2(m+n))$~\cite{golub2013matrix}. \\

  \textbf{Approximation by  sparse DMD.}\label{sec:sparse}  {\it Jovanovic et al.} also propose  in \cite{Jovanovic12} a two-stage approach which consists in searching a $k$  terms approximation  of the  EVD of an unconstrained projected  DMD, \ie the EVD of  $U_{\AAA}\tilde Y \Sigma_{\AAA}^{\dagger}U_{\AAA}^\intercal$.  The optimal truncated EVD is efficiently computed solving   a relaxed convex optimization problem taking advantage of the EVD of $\tilde Y \Sigma_{\AAA}^{\dagger} \in \Rr^{m \times m}$. The overall complexity of this procedure scales as  $\mathcal{O}(m^2(m+n))$.  \\

    \textbf{Approximation by  total-least-square DMD.}\label{sec:tls} This approximation is proposed by {\it Hemati et al.}~\cite{hemati2017biasing}.  Using the projector  $\mathbb{P}_{\mathbf{K}^\intercal,k} ={V}_{\mathbf{K},k}({V}_{\mathbf{K},k})^\intercal$ where the columns of ${V}_{\mathbf{K},k} \in \mathbb{R}^{m \times k}$ are the right singular vectors  associated  to the  $k$ largest  singular values of  matrix  $\mathbf{K}  =\begin{bmatrix} \AAA^\intercal & \BBB^\intercal \end{bmatrix}^\intercal\in \Rr^{2n \times m}$,
%
% The approximation introduced in {\it Hemati et al.}  can be formulated as the solution of the following unconstrained convex optimization problem 
%\begin{align}\label{eq:probHemati}
%\argmin_{A\in \Rr^{n \times n}} \| \BBB' - A \AAA' \|^2_F,
%\end{align}
%where $\AAA'= \AAA \mathbb{P}^k_{\mathbf{K}^\intercal} $ and $ \BBB'= \BBB\mathbb{P}^k_{\mathbf{K}^\intercal} $. 
the total-least-square DMD (TLS DMD)  approximation takes the form of
\begin{align}\label{eq:probHematiSol}
A_k^\star\approx \BBB\mathbb{P}_{\mathbf{K}^\intercal,k} \AAA^\dagger.
\end{align}
%may constitute an approximation of the solution of the  problem of interest, although
%the unconstrained problem~\eqref{eq:probHemati} is  intrinsically different from the low-rank approximation problem  \eqref{eq:prob1}. Indeed, we will see in the numerical simulations in Section~\ref{sec:numEval}  that the solution \eqref{eq:probHematiSol} of problem~\eqref{eq:probHemati}  is in certain settings an accurate approximation of  $A_k^\star$.
This method relies on the SVD of   $\mathbf{K}\in \Rr^{2n \times m}$ and $\XXX \in \Rr^{n \times m}$ and has thus a complexity of $\mathcal{O}(m^2(m+n))$. \\

   \textbf{Approximation by convex relaxation.}\label{sec:convex} 
 Some works propose to approximate~\eqref{eq:prob1} by a  regularized version of the unconstrained  problem~\eqref{eq:prob1_unconst}, using Tikhonov penalization \cite{li2017extended}  or  penalization  enforcing structured sparsity \cite{2017arXiv170806850Y}. However, these  choices of regularizers do not guarantee in general  that the solution is low-rank. In contrast,  the  solution of \eqref{eq:prob1} may under certain theoretical conditions~\cite{lee2010admira,jain2010guaranteed}  be recovered by  the following quadratic  program   
%The solution of \eqref{eq:prob1} may also be approximated 
\begin{align}\label{eq:probConvexRelas} 
 A_k^\star &\approx\argmin_{A\in \Rr^{n \times n}} \|\BBB -A \AAA \|^2_F+ \alpha_k \| A \|_{*},\nonumber \\
 &=  \argmin_{A\in \Rr^{n \times n}} \min_{B\in \Rr^{n \times n}} \|\BBB -A \AAA \|^2_F+ \alpha_k \| B \|_{*} \quad \textrm{s.t.} \quad A=B%\mbox{\quad$\forall\, s\in\Sc,t\in\Tc$},
\end{align} 
where $\| \cdot \|_{*}$  refers to the nuclear norm (or trace norm) of the matrix, \ie  the sum of its singular values.
In optimization problem~\eqref{eq:probConvexRelas}, $\alpha_k \in \Rr_+$ represents an appropriate regularization parameter determining the rank $k$ of the solution. 
%Particularized to problem \eqref{eq:prob1}, the theoretical results of , which are  related to the recovery of low-rank matrices in a noisy setting, provide a characterization of the conditions under which  the optimal solution $A_k^\star $  
Program~\eqref{eq:probConvexRelas} is a convex optimization problem \cite{mishra2013low} which can be efficiently solved using modern optimization techniques, such as the alternate directions of multipliers method (ADMM)~\cite{bertsekas1995nonlinear}. 
%Given $\tilde A_k^\star$, we can design a reduced model \eqref{eq:model_koopman_approx}  taking the form of the $k$-dimensional recursion \eqref{eq:genericROM} %. % Indeed, we introduce the adjoint of the pseudo-inverse of  ${U}_{\ZZZ,k}$  given by  $\G=\hat involving the projected variable  $z_t^i=  \R^\dagger \tilde x^i_t$:
%where we have set  $P=U_{\tilde A_k^\star}$ and ${Q}=V_{\tilde A_k^\star}\Sigma_{\tilde A_k^\star}$. Moreover, 
%Given $\tilde A_k^\star$, we can approximate low-rank DMD modes by the eigenvectors of $\tilde A_k^\star$ and deduce  amplitudes  from the associated eigenvalues according to~\eqref{eq:koopman2}. 
The algorithms solving \eqref{eq:probConvexRelas} typically  involve per iteration a complexity  of  $\mathcal{O}(m(m^2+n^2))$. 
\vspace{-0.15cm}
\subsection{Factorizations of  Approximations of   $ A_k^\star$}\label{sec:etatArt2}

In this section, we provide an overview of some state-of-the-art methods to compute %SVD or EVD factorizations 
% In this section, we  overview state-of-the-art methods to compute 
factorizations of the form of~\eqref{eq:solFactor} or~\eqref{eq:factorEVD} 
for the approximations of $ A_k^\star$ presented above.

%This approach presupposes the knowledge of a factorization of the form 
%\begin{align}\label{eq:solFactor}
%\hat A_k=PQ^\intercal \quad \textrm{with}\quad P,Q \in \Rr^{n \times r},
%\end{align}
%where $r \le n$. Assuming this structure, trajectories of \eqref{eq:model_koopman_approx} are   fully determined    by 
% the $r$-dimensional recursion~\eqref{eq:genericROM0} with $R=\R^\dagger$, $L=\R$ and $S= \Q^\intercal  \R$. According to Section \ref{sec:context}, the on-line complexity to compute this recursion is  $\mathcal{O}(r^2T+rn)$. 
We first note that a brute-force computation of the SVD or EVD of a matrix in $\Rr^{n \times n}$ leads in general to a prohibitive computational cost since it requires a  complexity of  $\mathcal{O}(n^3)$. Hopefully,  the factorizations \eqref{eq:solFactor} or~\eqref{eq:factorEVD}  are computable with a  complexity of  $\mathcal{O}(m^2(m+n)) $,  in most cases mentioned above.
 % with    $r=\textrm{rank}({\hat A_k})\le k \le m$. 
 
In particular,  in the case of low-rank projected DMD, a straightforward  factorization of the form of~\eqref{eq:solFactor}  is $P=U_\AAA$ and $Q^\intercal = \tilde Y_k \Sigma_{\AAA}^{\dagger}U_{\AAA}^\intercal$. In the case of  sparse DMD, the latter factorization holds by substituting   $\tilde Y_k$ by the ``sparse'' approximation of $\tilde Y$. Another straightforward  factorization of the form of~\eqref{eq:solFactor} is  intrinsic to the ADMM procedure, which uses an SVD  to compute the regularized solution.

Concerning EVD factorization,  in the case of the truncated approach, {\it Tu et al.} propose an algorithm scaling in   $\mathcal{O}(m^2(m+n))$~\cite{Tu2014391}. 
 In the context of low-rank  projected DMD or sparse DMD, {\it Jovanovic et al.} propose a  procedure of analogous  complexity, which approximates the first $m$ eigenvectors, and then estimate the related eigenvalues by solving a convex optimization problem \cite{Jovanovic12}. In the case of TLS DMD, the diagonalization of a certain matrix in $\Rr^{m \times m}$ suffices to obtain the sought EVD factorization. \\ % implies the same computation of matrix $R$  followed by a sparse convex minimization problem to obtain amplitudes $\nu_{i,t}$ 
  %No efficient algorithm is provided in the literature for computing matrices $R$, $L$ and $S$ in the case where$\hat A_k$ is obtained by SVP as described in Section~\ref{sec:IHT}.  However,   we  expect from matrix analysis that the EVD of $A^\star_k$ can be efficiently computed taking advantage of the $k$-term SVD of $\hat A_k$ given by the SVP algorithm \cite{golub2013matrix}. 

In summary,  on the one hand, we saw in Section~\ref{sec:approxLDMD} that all existing algorithms compute in general sub-optimal solutions of problem \eqref{eq:prob1}. On the other hand,  the literature does not provide a ``turnkey" algorithm for factorizing the optimal solution  $A^\star_k$ in the form of \eqref{eq:solFactor} or \eqref{eq:factorEVD}, with a reduced complexity of $\mathcal{O}(m^2(m+n)) $.
 In the next section, we show  how to compute an optimal solution $A^\star_k$ and  compute efficiently its factorization.\vspace{-0.15cm}
%Indeed the low-rank property implies that the matrix rows and columns span a specific low-dimensional subspaces. 
%I
\section{ The Proposed Approach}\label{sec:contrib}

In this section, we provide a closed-form solution to problem~\eqref{eq:prob1}.  Algorithms are then proposed to compute and factorize this solution in the form of~\eqref{eq:solFactor} or~\eqref{eq:factorEVD}. \vspace{-0.15cm}

\subsection{Closed-Form Solution to~\eqref{eq:prob1}}\label{sec:closedSol}

 Let the columns of matrix  ${U}_{\ZZZ,k} =\begin{pmatrix} u^1_\ZZZ& \cdots &u^k_\ZZZ  \end{pmatrix}\in \Rr^{n \times k}$ be the  left singular vectors $\{u_\ZZZ^i\}_{i=1}^k$ associated  to the  $k$ largest  singular values of  matrix  \begin{align}\label{eq:defZZZ} \ZZZ= % \BBB \AAA^\dagger\AAA=
 \BBB \mathbb{P}_ {\AAA^\intercal}\in \Rr^{n \times m},\end{align}  where we recall that $ \mathbb{P}_ {\AAA^\intercal}={V}_\AAA{V}_\AAA^\intercal$ and consider  the projector 
\begin{align}\label{eq:hatP}
\mathbb{P}_{\ZZZ,k}={U}_{\ZZZ,k} {{U}_{\ZZZ,k}}^\intercal.
\end{align}
Matrix \eqref{eq:hatP} appears in the closed-form solution of~\eqref{eq:prob1}, as shown in  the following  theorem. We detail the proof  in Appendix \ref{app:Theorem}.

\noindent
\begin{theorem}\label{prop22}
%Assume that  
%		%\begin{align}\label{eq:assumptions}
%		$\textrm{rank}(\AAA)\ge k.$ %\quad  \textrm{and}\quad \textrm{rank}(\BBB)\ge k.
%		%\end{align}
%Then, 
Problem~\eqref{eq:prob1}  admits the following solution 
% \begin{itemise}
% \item   ${U}_{\ZZZ,k}=(\hat p_1 \cdots \hat p_k)$, whose columns  are real orthonormal  eigenvectors  associated  to the  $k$ largest  eigenvalues  of  the symetric matrix  $\BBB\AAA^\intercal (\AAA\AAA^\intercal )^{\dagger}\AAA\BBB^\intercal ,$
% \item   ${W}=(\hat q_1 \cdots \hat q_k)$,  with $\hat q_j=(\AAA\AAA^\intercal )^{\dagger}\AAA\BBB^\intercal \hat p_j$, for $j=1,...,k$.
%   \end{itemise}  
%   Moreover, the matrix 
\begin{align}\label{eq:Sol}
A_k^\star= \mathbb{P}_{\ZZZ,k}   \BBB \AAA^{\dagger}.
\end{align} % is a  solution  of~\eqref{eq:prob1}, and the unique matrix solution with rows belonging to $\textrm{Im}(\AAA)$. %, up to a permutation of the columns of ${U}_{\ZZZ,k}$ and ${W}$.
Moreover, the optimal approximation error can be expressed as
\begin{align}\label{eq:errorEstim}
 \|\BBB -A_k^\star \AAA \|^2_F = \sum_{i=k+1}^m \sigma_{\ZZZ,i}^2 +\| \BBB (I_m-\mathbb{P}_{\AAA^{\intercal}})\|_F^2.
\end{align}
  \end{theorem}

 In words, Theorem~\ref{prop22} shows that problem~\eqref{eq:prob1} is simply solved by computing the orthogonal projection of  the  solution of the unconstrained  problem~\eqref{eq:prob1_unconst},  onto the subspace spanned by the first $k$    left singular vectors  of  $ \ZZZ.$    The $\ell_2$-norm of the error is simply expressed in terms of  the singular values of $\ZZZ$, and  the square norm of the projection of the rows of $\BBB$ onto the orthogonal of the image of $\AAA^\intercal$.
If $\AAA$ is full {row-}rank,  we then obtain  the simplifications $\mathbb{P}_{\AAA^\intercal}=I_m$ and $\ZZZ=\BBB$. In this case, the second term in the right-hand side of~\eqref{eq:errorEstim} vanishes and the  approximation error reduces to $ \|\BBB -A_k^\star \AAA \|^2_F = \sum_{i=k+1}^m \sigma_{\BBB,i}^2$. The latter error is independent of matrix $\AAA$ and is simply the sum of the square of the $m-k$ smallest singular values of $\BBB$. This error also corresponds to the optimal error for the approximation $\BBB$ by a matrix of rank at most $k$ in the Frobenius norm \cite{eckart1936approximation}.  
%Besides, $r=\textrm{rank}(A_k^\star)$  can be smaller than $k$. Indeed, by the Sylvester's theorem \cite{Horn12} we have  that
%\begin{align*}
%\textrm{rank}(A_k^\star)&\le\min(\textrm{rank}(\mathbb{P}_{\ZZZ,k} ), \textrm{rank}(\BBB\XXX^\dagger))  \le  \textrm{rank}(\BBB\XXX^\dagger)\\
%&\le \min(\textrm{rank}(\BBB), \textrm{rank}(\AAA^\dagger))= \min(\textrm{rank}(\BBB), \textrm{rank}(\AAA)),
%\end{align*} 
%which shows that $r<k$ if $\textrm{rank}(\XXX)$ or $\textrm{rank}(\YYY)$ is smaller than $k$, but also if $\textrm{rank}(\BBB\XXX^\dagger) < k$. 

It is worth mentioning that  we propose in~\cite{HeasHerzet18Maps} a generalization of Theorem~\ref{prop22} to separable infinite-dimensional Hilbert spaces. This generalization  characterizes the solution of  low-rank approximations in reproducing kernel Hilbert spaces (where $n=\infty$) at the core of  kernel-based DMD \cite{HeasIcassp2020,williams2014kernel}, and characterizes the solution of the DMD counterpart (where $m=\infty$) to  the continuous POD problem presented in \cite[Theorem 6.2]{quarteroni2015reduced}. \vspace{-0.15cm}

%The latter condition does not necessarily  imply the former.

\subsection{Algorithm Evaluating $A_k^\star$}\vspace{-0.35cm}

\begin{algorithm}[!h]
\begin{algorithmic}[0]
\State \textbf{inputs}: $(\XXX,\YYY).$%, $\delta$ \remCH{changer tolerence par "stopping criterion"}
\State 1)  Compute the SVD of $\AAA= V_{\AAA} \Sigma^\dagger_{\AAA} U_{\AAA}^\intercal  $ 
\State 2)  Compute $\mathbf{Z}=\BBB V_{\AAA}\Sigma_{\AAA}\Sigma_{\AAA}^\dagger V_{\AAA}^\intercal$.% and $\AAA^\dagger= V_{\AAA} \Sigma_{\AAA}^{\dagger}U_{\AAA}^\intercal $.
\State 3) Compute the SVD of $\mathbf{Z}$ to obtain the projector $\mathbb{P}_{\ZZZ,k}$.
%\State 4) Compute  the  first $k$ columns of $V_\ZZZ$ and  $\Sigma_\ZZZ^{2}$, \ie  the first  $k$  eigenvectors/eigenvalues  of $\mathbf{Z}^\intercal  \mathbf{Z} $.
%\State 5) Compute the columns of  ${U}_{\ZZZ,k}$ defined as the  first $k$ columns of matrix $  \mathbf{Z} V_\ZZZ \, \Sigma_\ZZZ^{\dagger}$.%,  by making the identification  $Z=\BBB V_{\AAA}  $ in Remark \ref{rem:1}. 
 %\State 4) Compute  the columns of  ${W}$, where the $j$-th column for $j=1,...,k$ is given by  $$\hat q_j=  U_{\AAA} {\Sigma_{\AAA}}^{-1}V_{\AAA}^\intercal \BBB^\intercal \hat p_j.$$
%\remCH{Cette derniere operation peut sans doute encore etre simplifee en utilizant l'expression de la SVD de $\BBB$.}
\State 4) Compute $A_k^\star= \mathbb{P}_{\ZZZ,k}   \BBB  V_{\AAA} \Sigma^\dagger_{\AAA} U_{\AAA}^\intercal  $.
\State \textbf{output}: $A_k^\star$. % $({U}_{\ZZZ,k} ,\BBB, U_{\AAA}, \Sigma_{\AAA}, V_{\AAA})  $%, $\delta$ \remCH{changer tolerence par "stopping criterion"}
\end{algorithmic}
\caption{Computation of $A_k^\star$, a solution of~\eqref{eq:prob1} \label{algo:1}}
\end{algorithm}
The design of an algorithm computing   the solution~\eqref{eq:Sol} is straightforward: evaluating $A_k^\star$ consists in making a product of easily-computable matrices.   The proposed procedure is summarized in Algorithm~\ref{algo:1}.

Steps 1) to 3) of  Algorithm~\ref{algo:1} implies the computation of the SVD of  matrices $\AAA,\ZZZ \in \Rr^{n \times m }$, and matrix multiplications involving $m^2$ vector products in $\Rr^n$ or $\Rr^m$. %and $k$ multiplications of a matrix in $\Rr^{n \times m }$ by a vector in order to obtain the $k$ first columns of $ U_\ZZZ$. 
The  complexity  of these  first three  steps is therefore  $\mathcal{O}(m^2(m+n))$. 
 Computing explicitly each entry of  $A_k^\star \in \Rr^{n\times n}$ in  step 4) of Algorithm~\ref{algo:1}  then requires a  complexity of   $\mathcal{O}(n^2k)$, which is prohibitive for large $n$. However, as detailed in the next section, this last step is not necessary to   factorize the optimal solution  $A^\star_k$ in the form of \eqref{eq:solFactor} or \eqref{eq:factorEVD}.  \vspace{-0.15cm}

%For  problems with large $m$, the cubic complexity to diagonalize matrix $ \ZZZ^\intercal \ZZZ$ could be lowered by using a power iteration method able to approximate a leading subset of eigenvalues and eigenvectors. 

%\remCH{NB: la complexite de l'algo tu proposes scale en $\mathcal{O}(m^3)$ (idem que POD). Quelle est la complexite des autres methodes?} 

\subsection{Algorithms Factorizing $A_k^\star$}
Given the closed-form solution~\eqref{eq:Sol}, we  present in what follows how to compute from $\AAA$ and $\BBB$ a factorization of the optimal solution  $A^\star_k$ in the form of \eqref{eq:solFactor} or \eqref{eq:factorEVD}. We will need 
  matrix
\begin{align}\label{eq:defhatQ}
 {W} =({{U}_{\ZZZ,k}}^\intercal  \BBB \AAA^\dagger)^\intercal \in \Rr^{n \times k}.%\quad \textrm{and}\quad \hat S= {U}_{\ZZZ,k}{W}^\intercal{U}_{\ZZZ,k} \in \Rr^{n \times k}.
 \end{align} \vspace{-0.15cm}

\textbf{Factorization of the form of \eqref{eq:solFactor}.} 
%\begin{algorithm}[t]
%\begin{algorithmic}[0]
%\State \textbf{inputs}:  $(\XXX,\YYY).$
%\State 1) Compute ${U}_{\ZZZ,k}$ performing step 1 to 3 of Algorithm~\ref{algo:1}.
%\State 2) Compute ${W}$ using~\eqref{eq:defhatQ}.
%\State \textbf{outputs}:    $L={U}_{\ZZZ,k}$, $R={{U}_{\ZZZ,k}}^\intercal$, $S= {W}^\intercal  {U}_{\ZZZ,k}$. %, $\delta$ \remCH{changer tolerence par "stopping criterion"}
%\end{algorithmic}
%\caption{SVD-based factorization of $A^\star_k$  \label{algo:2bis}}
%\end{algorithm}
By performing the first three steps  of Algorithm~\ref{algo:1} and then making the identifications  $\R={U}_{\ZZZ,k}$ and $\Q={W}$, we obtain a factorization of $A_k^\star$ of the form of \eqref{eq:solFactor}. As mentioned in the introduction, trajectories of \eqref{eq:model_koopman_approx} can then be computed with system~\eqref{eq:genericROM0}  setting $R={U}_{\ZZZ,k}$, $L=W$ and $S=  W^\intercal  {U}_{\ZZZ,k}$.
The method relies on the  first three steps  of Algorithm~\ref{algo:1} and on the computation of matrix ${W} $. The three steps in Algorithm~\ref{algo:1} imply a complexity of $\mathcal{O}(m^2(m+n))$ while the  computation of ${W} $ requires a complexity of $\mathcal{O}(n k^2)$. Since $k \le m$, the off-line complexity to build  the  factorization~\eqref{eq:solFactor}  from  $\AAA$ and $\BBB$ scales as $\mathcal{O}(m^2(m+n))$, which is the same order of complexity as the procedures described in Section~\ref{sec:stateArt}.  \\
%for state-of-the-art algorithms.  \vspace{-0.15cm}

\begin{algorithm}[t]
\begin{algorithmic}[0]
\State \textbf{inputs}:  $(\XXX,\YYY).$
\State 1) Compute step 1 to 3 of Algorithm~\ref{algo:1} and  use~\eqref{eq:defhatQ} to obtain ${W}$.
%\State 2) Compute the matrices  $$D^r_k={W}^\intercal {U}_{\ZZZ,k}  \in \Rr^{k \times k} \quad \textrm{and}\quad D^\ell_k=(D^r_k)^\intercal.$$
%=U_{\hat\Q}\Sigma_{\hat\Q} V_{\hat\Q}^\intercal $.%\ie and  the SVD of matrix $\AAA$ %, we obtain are able to compute ${W}\in \Rr^{m \times m}$ satisfying 
%\begin{align*}
%\AAA&= U_a \,\begin{pmatrix}\textrm{diag}({\sigma_a}^{\frac{1}{2}})\\ 0_{n-\ell}\end{pmatrix}    V_a ^\intercal .
%\end{align*}
%\State 2)  Define matrix   $D^r_k=U_{\hat\Q}^\intercal {U}_{\ZZZ,k} V_{\hat\Q}\Sigma_{\hat\Q} \in \Rr^{m\times m}$.
\State 2) Let $r=\textrm{rank}(A^\star_k)$ and solve for $ i=1, \ldots,r$ the eigen-equations $$ ({W}^\intercal {U}_{\ZZZ,k}) w^r_i = \lambda_i w^r_i\quad \textrm{and}\quad ({{U}_{\ZZZ,k}}^\intercal {W}) w^\ell_i =  \lambda_i w^\ell_i,$$ where  $w^r_i, w^\ell_i \in \Cr^k$  %denote  the right and left eigenvectors   of ${W}^\intercal {U}_{\ZZZ,k} \in \Rr^{k \times k}$.  The
and $\lambda_i\in \Cr$ such that $|\lambda_{i+1}| \ge |\lambda_i |$. % denote the non-zero eigenvalues of ${W}^\intercal {U}_{\ZZZ,k}$ and ${{U}_{\ZZZ,k}}^\intercal {W}$. 

%\State 3) The $i$-th DMD mode corresponding to the eigenvalue $\lambda_i$ is then given by 
%$$ \phi_i=.$$
\State 3) Compute for $i=1,\ldots,r$ the right  and  left eigenvectors  
\begin{align}\label{eq:defEigenvectors}
  \zeta_i=  {U}_{\ZZZ,k}   w^r_i\quad \textrm{and}\quad   {\xi_i}=  {W} w^\ell_i.
  \end{align} 
  \State 4) Rescale the ${\xi_i}$'s so that $  {\xi_i}^T \zeta_i =1$. 
\State \textbf{outputs}:   $L=(\xi_1\cdots \xi_r)$,  $R=(\zeta_1\cdots \zeta_r)$,    $S=\textrm{diag}(\lambda_1,\cdots, \lambda_r).$ % $\zeta_i$ and  $\nu_{i,t}=\lambda_i^{t-1} \xi_i^\intercal \theta$ for  $i=1,\cdots,k'$ and any $t>1$. %, $\delta$ \remCH{changer tolerence par "stopping criterion"}
\end{algorithmic}
\caption{EVD of $A^\star_k$ or low-rank DMD  \label{algo:2}}
\end{algorithm}

\textbf{Factorization of the form of  \eqref{eq:factorEVD}.}\label{sec:EVDAstar} 
%Although it is linear in $n$, running this algorithm becomes very expensive or even prohibitive in the case where$n \gg p$. This situations typically occurs  since the Koopman operator approximation is in general accurate for $n$ tending to infinity \cite{williams2015data}. This remark motivates the design of a second algorithm dedicated to the case where$n \gg p$. By exploiting the kernel trick and avoiding any computation in $\Rr^n$, this alternative algorithm  estimates the parameters  of  reduced model~\eqref{eq:koopman1}  with a complexity  of  $\mathcal{O}(m^2(m+p))$.  
According to the previous factorization of the form of \eqref{eq:solFactor}, $A^\star_k$ is  the product of matrix ${U}_{\ZZZ,k}$ in $\Rr^{n \times k}$ with matrix ${W}^\intercal$ in $\Rr^{k \times n}$. Therefore, using  standard matrix analysis, we expect the eigenvectors of $A^\star_k$ to belong to a $k$-dimensional  subspace  \cite{golub2013matrix}. As shown in  the next proposition,  the non-zero eigenvalues of $A^\star_k$  are obtained by EVD of certain matrices in $\Rr^{k \times k}$. The proof of this proposition  is given in Appendix~\ref{app:prop}.

 \begin{proposition}\label{rem:2}
 Assume $A^\star_k$ is diagonalizable.   The elements of  $\{\zeta_i, \xi_i, \lambda_i\}_{i=1}^{\mathrm{rank}(A^\star_k)}$ generated  by Algorithm~\ref{algo:2}  are  the  right eigenvectors, the left eigenvectors and the eigenvalues  of the economy size EVD of  $A^\star_k$. % Moreover they satisfy  $\xi_i ^\intercal A^\star_k \zeta_i =\lambda_i$.\\
 \end{proposition}

In words, Proposition~\ref{rem:2} shows that Algorithm~\ref{algo:2} computes  the  EVD   of  $A^\star_k$ by  diagonalizing  two  matrices in  $\Rr^{k\times k}$. 
 % $$D^r_k=U_{\hat\Q}^\intercal {U}_{\ZZZ,k} V_{\hat\Q}\Sigma_{\hat\Q} \in \Rr^{m\times m}.$$  
 The complexity  to build  this  EVD  from snapshots $\XXX$ and $\YYY$ is  $\mathcal{O}(m^2(m+n))$. More precisely, as mentioned previously, performing the first three  steps of Algorithm~\ref{algo:1} (\ie step 1) of Algorithm~\ref{algo:2}) requires a number of operations scaling as $\mathcal{O}(m^2(m+n))$;
 % we saw that step 1), \ie   the  first three  steps of Algorithm~\ref{algo:1}, is run  with a complexity of $\mathcal{O}(m^2(m+n))$; 
  the complexity of step 2) is $\mathcal{O}(k^3)$ since it  performs the EVDs of  $k \times k$ matrices; 
  % $({W}^\intercal {U}_{\ZZZ,k}) \in \Rr^{k\times k}$ and its transpose; 
  step 3) involves $r\times n$ vector products in $\Rr^m$ while step 4) involves $r$ vector products in $\Rr^n$, with $r \le k \le m$. 
  {Overall, the complexity of Algorithm~\ref{algo:2} is dominated by step 1) and the EVD of $A^\star_k$ can be evaluated with a computational cost of the order of $\mathcal{O}(m^2(m+n))$.}\vspace{-0.25cm}

\section{Numerical Evaluation}\label{sec:numEval} 

In what follows, we evaluate  five different trajectory approximations $ \tilde x_{t}(\theta)$ obtained by reduced model of the form of~\eqref{eq:koopman1}, obtained by  EVD factorizations of  different low-rank matrix approximations $\hat A_k$. The assessed low-rank matrix factorizations are listed below.
\begin{itemize}
\item  \textbf{Optimal approximation}: the EVD of the optimal solution $A_k^\star$, provided by  Algorithm~\ref{algo:2}. \vspace{-0.cm}
\item  \textbf{Approximation by truncated DMD} \cite{Tu2014391}: 
 the $k$-th order truncation of the EVD of  the unconstrained problem solution $A^\star_m$ given in~\eqref{eq:exactDMD}.\vspace{-0.cm}
\item  \textbf{Approximation by low-rank projected DMD} \cite{Jovanovic12}:   the EVD of the $k$-th order approximation~\eqref{eq:projDMD}.\footnote{We do not  evaluate the sparse DMD approach since  the error norm  induced by this method will always be greater than the one induced by  low-rank  projected DMD, see details in~\cite{HeasHerzet17}.} \vspace{-0.cm}
\item  \textbf{Approximation by TLS DMD} \cite{hemati2017biasing}:  the  EVD of~\eqref{eq:probHematiSol}.  \vspace{-0.cm}
\item  \textbf{Approximation by convex relaxation}:  the  EVD of the solution~\eqref{eq:probConvexRelas}, the latter is computed by an ADMM procedure, where the regularization parameter $\alpha_k$ is adjusted to obtain a rank equal to $k$.  \vspace{-0.cm}
\end{itemize}

Rather than evaluating the error norm of the approximation, \ie the cost of the target problem~\eqref{eq:target}, we are interested in the ability of the different algorithms to minimize the cost of the proxy~\eqref{eq:prob1} for this problem. Therefore, the performance is measured in terms of the normalized  reconstruction error norm ${\|\BBB-\hat A_k \AAA\|_F}{\|\BBB\|_F^{-1}}$ as a function of~$k$.  Besides, in the analysis perspective adopted most often in the DMD literature \cite{hemati2017biasing,Jovanovic12,Schmid10,Tu2014391}, we are interested in evaluating the ability of the algorithms to compute accurately the EVD of $A^\star_k$. In particular, we  quantify  for a given~$k$ the deviation of the set of the $k$ largest estimated eigenvalues $\lambda(\hat A_k)\in \mathbb{C}^k$,   from the non-zero optimal ones $\lambda( A^\star_k)$ in terms of the normalized error norm ${\|\lambda( \hat A_k)-\lambda( A^\star_k)\|_2}{\|\lambda( A^\star_k)\|_2^{-1}}$.

We begin by evaluating the performance of the low-rank approximations on a different toy models in Section~\ref{sec:toy}. We then assess their performance for the reduction of a Rayleigh-B\'enard convective system~\cite{chandrasekhar2013hydrodynamic} in Section~\ref{sec:phys}. We finally evaluate the influence of noise on the estimation accuracy  in Section~\ref{sec:robust}.  \vspace{-0.75cm}

\subsection{ Synthetic Experiments with Toy Models}\label{sec:toy}

\begin{figure}[b!]
 \centering\vspace{-2.5cm}
\begin{tabular}{cc}
\includegraphics[width=0.525\columnwidth]{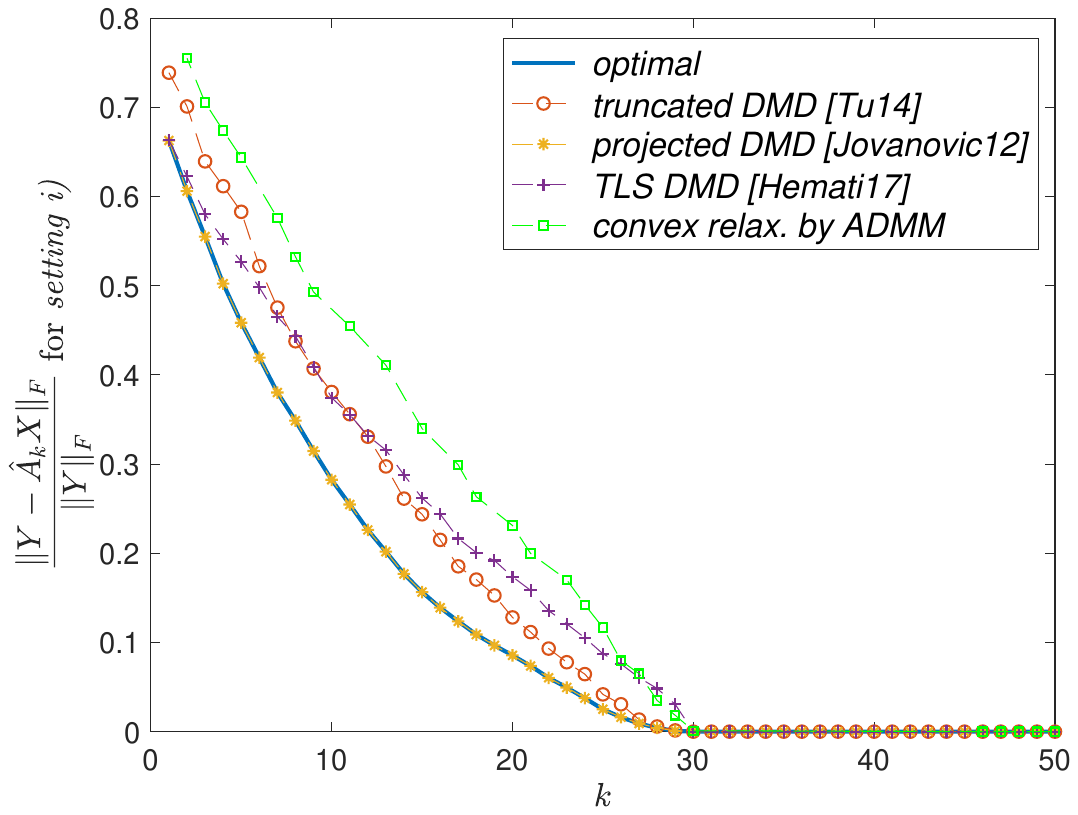}&\hspace{-1.cm}
\includegraphics[width=0.525\columnwidth]{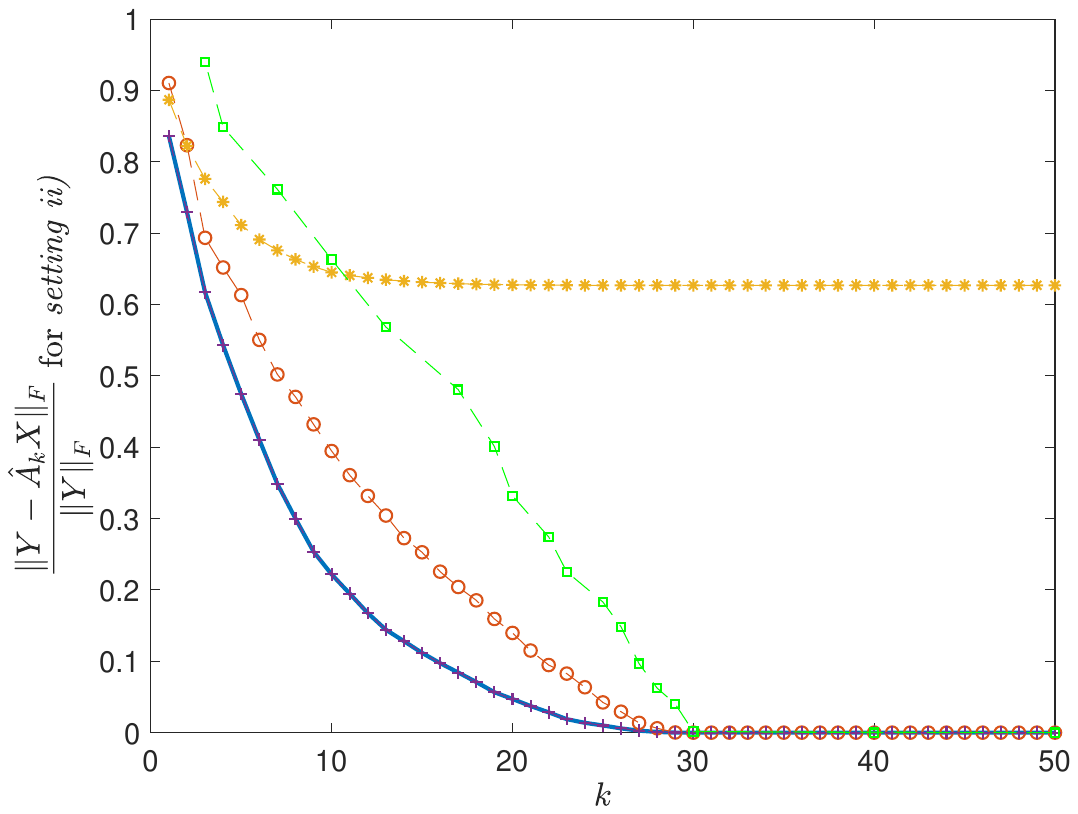}\vspace{-4.5cm}\\
\includegraphics[width=0.525\columnwidth]{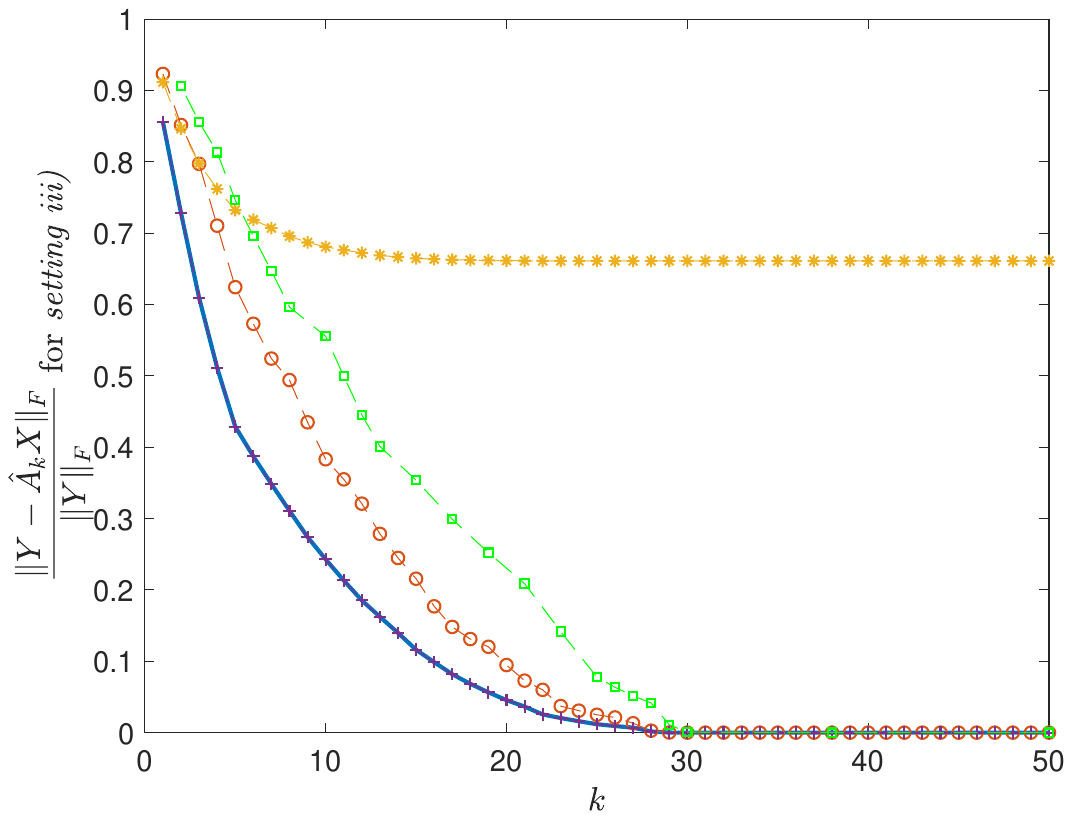}&\hspace{-1.cm}
\includegraphics[width=0.525\columnwidth]{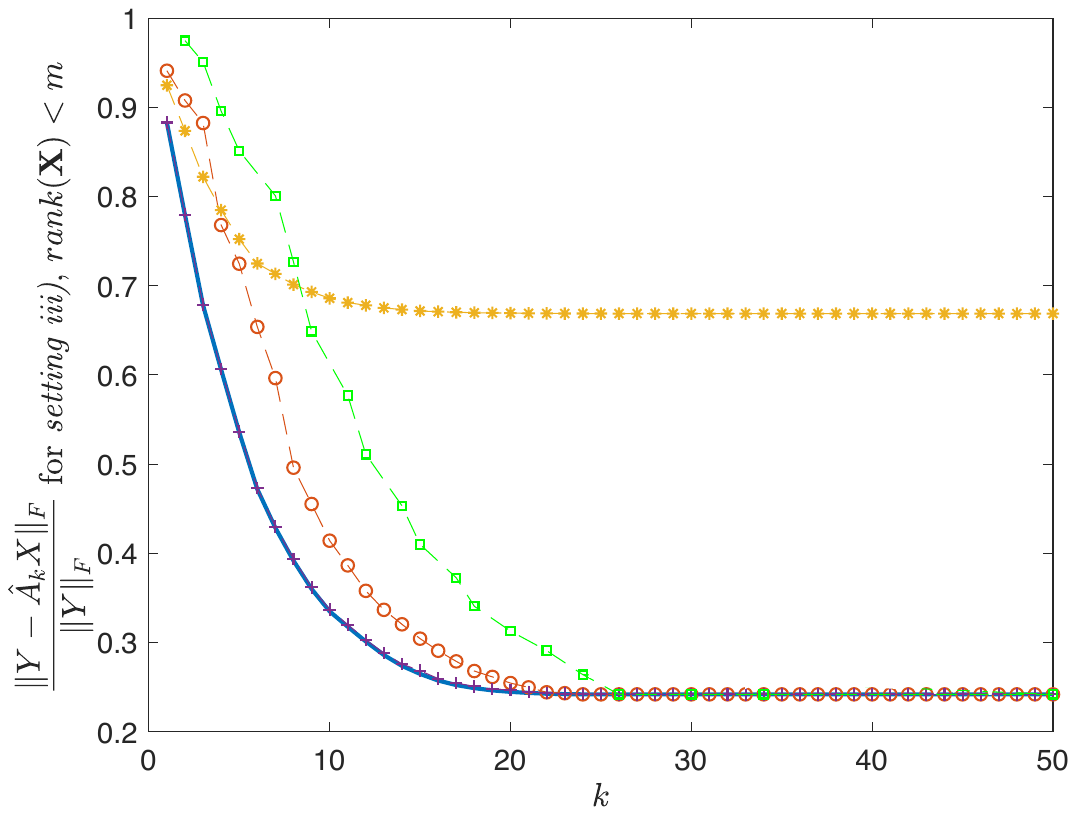}\vspace{-2.5cm}
\end{tabular}
\caption{\small   Reconstruction error norm  as a function of  $k$ for   {\it setting $i)$, $ii)$} and $iii)$ (for $rank(\XXX)=m$ and $rank(\XXX)=m-6$) using our optimal approximation or state-of-the-art approximations. See details in Section~\ref{sec:toy}.  \vspace{-0.5cm}\label{fig:1}}
\end{figure}

 \begin{figure}%[h!]
\centering\vspace{-2.cm}
\begin{tabular}{cc}
\includegraphics[width=0.525\columnwidth]{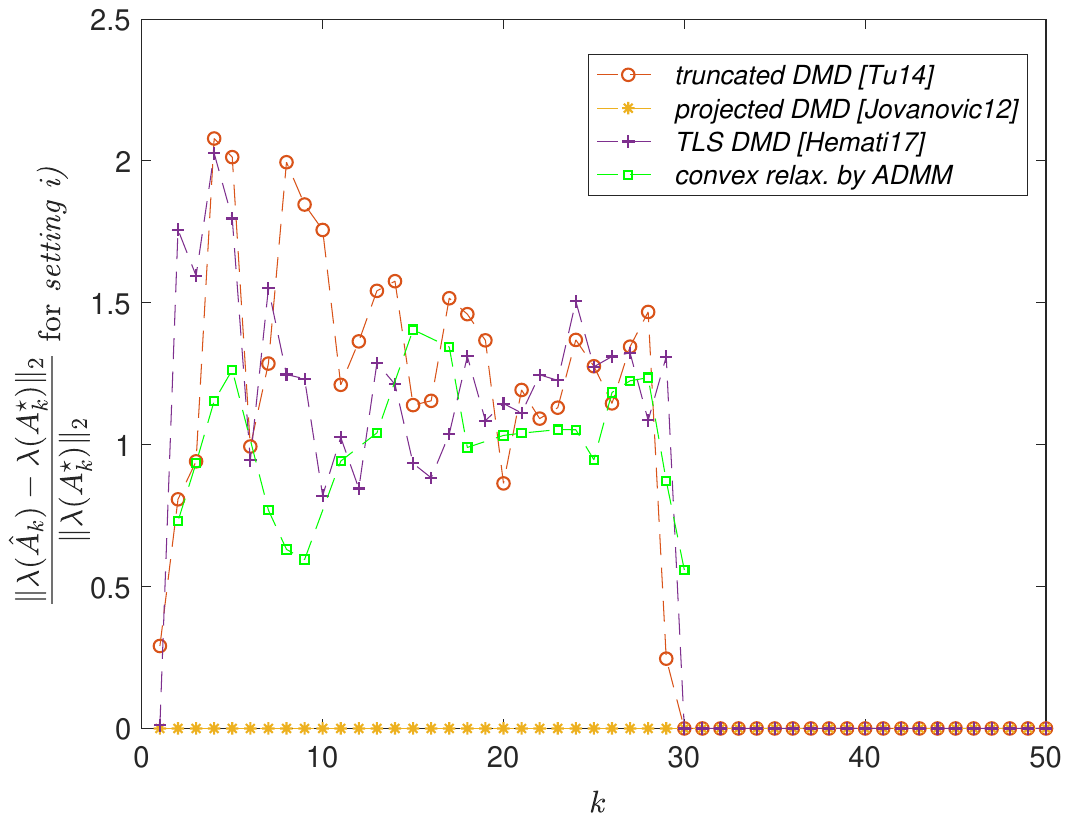}&\hspace{-1.cm}
\includegraphics[width=0.525\columnwidth]{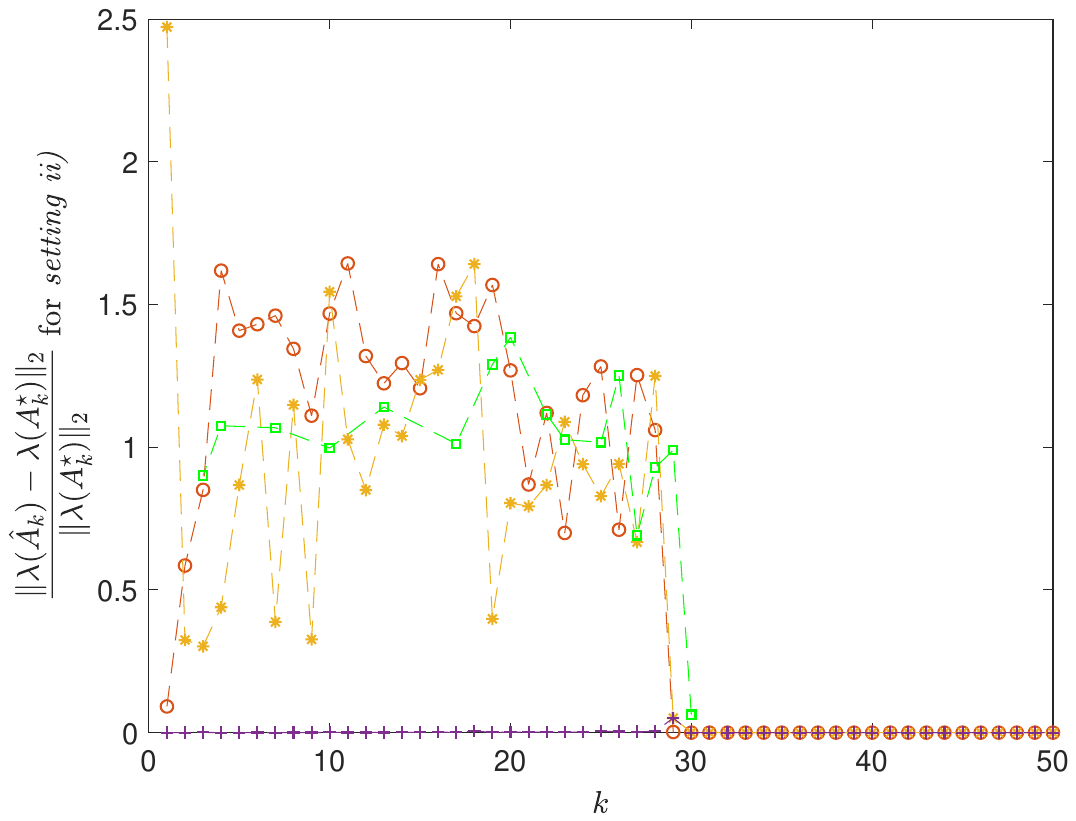}\vspace{-4.5cm}\\
\includegraphics[width=0.525\columnwidth]{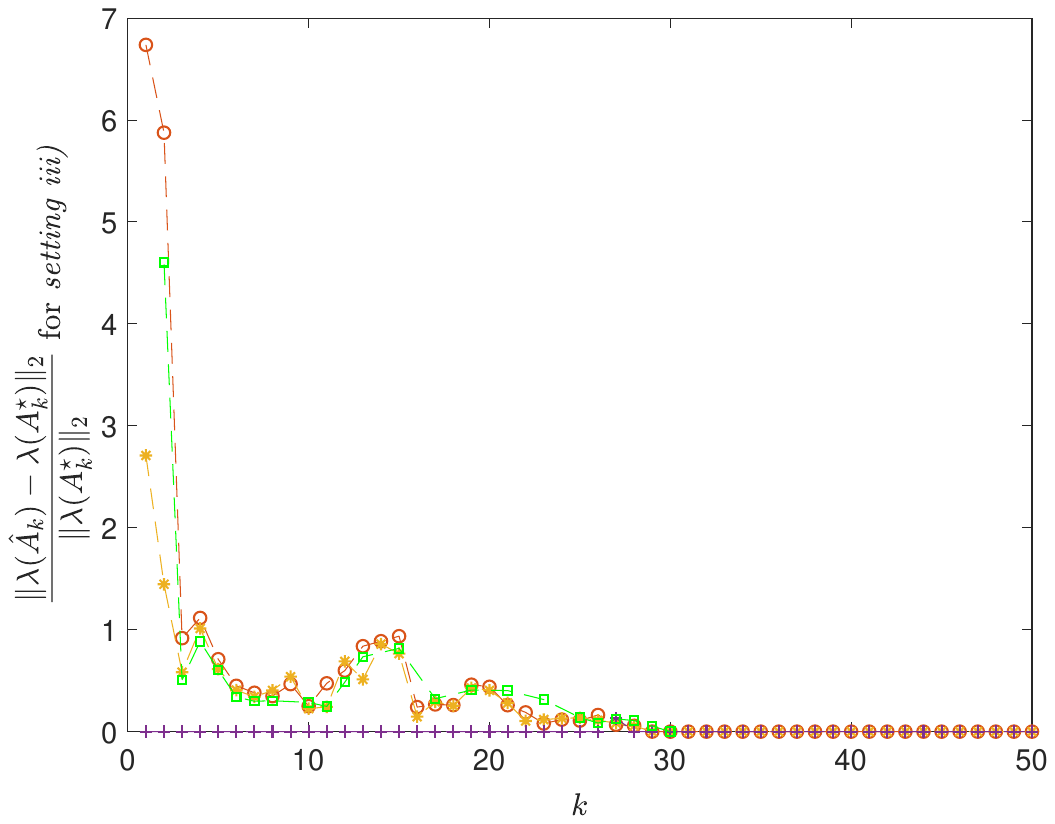}&\hspace{-1.cm}
\includegraphics[width=0.525\columnwidth]{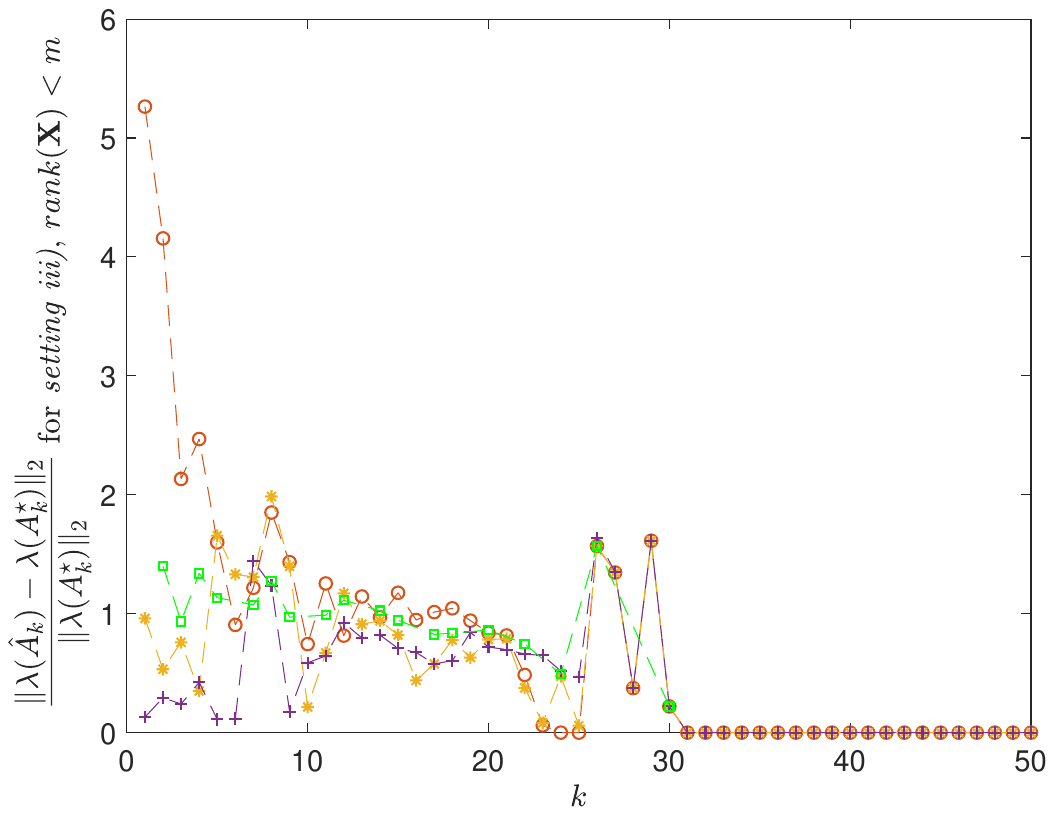}\vspace{-2.5cm}
\end{tabular}
\caption{\small   Eigenvalue error norm  as a function of  $k$ for   {\it setting $i)$, $ii)$} and $iii)$ (for $\mathrm{rank}(\XXX)=m$ and $\mathrm{rank}(\XXX)=m-6$) using our optimal approximation or state-of-the-art approximations. See details in Section~\ref{sec:toy}.  \vspace{-0.4cm}\label{fig:1eig}}
\end{figure}

We set $n=50$, $N=30$ and set the trajectory length in  \eqref{eq:model_init} to $T=2$.  Entries of matrices $\AAA$, \ie the initial condition $\theta$,  are such that $\AAA=\sum_{i=1}^{\textrm{rank}(\AAA)} \varphi_i \varphi_i^\intercal $ with $\textrm{rank}(\AAA) \le m$ and where the $\varphi_i$'s are $n$-dimensional  independent samples of the standard normal distribution.     %  drawn according to the $n$-dimensional standard normal distribution. 
Matrix $\BBB$ is then generated using model~\eqref{eq:model_init} %from a random initial condition $\theta$ 
and three different choices for $f_t$:   \vspace{-0.cm}
\begin{itemize}
\item  {\it setting {i)}}:\,\, $f_t(x_{t-1})=G x_{t-1}$, where $G$ is chosen such that $G \AAA \in \textrm{span}(\AAA)$,\vspace{-0.cm}
\item   {\it {setting ii)}}:\,\, $f_t(x_{t-1})=F x_{t-1}$,\vspace{-0.cm}
\item   {\it {setting iii)}}:  \,\,$f_t(x_{t-1})=F (x_{t-1}+x_{t-1}^3)$. \vspace{-0.cm}
\end{itemize}

Matrix $F$ introduced above is a random matrices of rank $m$ %\remCH{tu utilises deja la notation a) comme variable d'optimization precedemment.} 
defined as $F=\sum_{i=1}^{m} \varphi_i \varphi_i^\intercal $,  where the $\varphi_i$'s represents a set of $n$-dimensional  independent samples of the standard normal distribution. We define implicitly matrix $G$  of rank $m$ by drawing the parameters of the so-called companion matrix~\cite{Schmid10}  with  independent samples of the standard normal distribution. The notation $x_{t-1}^3$ denotes that each entry of vector $x_{t-1}$ has been raised to the power $3$. %The pseudo-inverse $\AAA^\dagger$  is computed from the  SVD of $\AAA$. %\remCH{pourquoi consideres tu uniquement des matrices definies positives? Ton systeme est-il stable (quel est le spectre des valeurs propres)?} 
The first setting is a linear system satisfying the sub-space assumption  $G \AAA \in \textrm{span}(\AAA)$, on which  low-rank  projected DMD relies. The two next settings  do not make this assumption and simulate respectively  linear and cubic dynamics.
The performance of the  five methods in terms of reconstruction and eigenvalue errors are displayed in Figure~\ref{fig:1} and \ref{fig:1eig}.

 %\remCH{ca fait un peu bizarre que tu annonces 3 setups et que tu n'en montres que 2.}
In accordance with our theoretical results, the proposed algorithm yields the smallest error norms in all the scenarios. Moreover, in accordance with Theorem~\ref{prop22}, as long as we have a full-rank matrix $\AAA$,  the optimal solution reaches a zero reconstruction error for $k \ge m$  with  $m=N(T-1)=30$ (we have considered $N=30$ snapshots and  $T=2$ successive states).  
% We remark that the results are in agreement.
The deterioration of the reconstruction error norm  for the approximation by truncated DMD shows that a two-stage approach  is sub-optimal. However, this deterioration is moderate in these toy experiments. %We mention that, although not displayed in the figure,  the  gap  with the optimal solution  becomes important for $k<r$ if we choose to truncate the  EVD\footnote{As pointed out previously, this alternative two-stage method is voluntarily not displayed  to lighten the presentation.} of $A^\star_m$ instead of its SVD.  
Moreover, the experiments show  that  the approximation by low-rank projected DMD  achieves the optimal performance  as long as  the  sub-space assumption  holds, \ie for {\it setting~i)}. If this assumption is not satisfied, \ie in {\it setting  ii)} and {\it iii)}, we observe a poor  performance of this projected approach for $k>10$. %The error norm with respect to $k$ is not even necerssarily a decreasing function  in the case where $rank(\XXX)<m$.  
On the contrary to low-rank  projected DMD, TLS DMD performs poorly in the case where $G \AAA \in \textrm{span}(\AAA)$, \ie for {\it setting~i)}, while it yields a quasi-optimal error norm in the other settings. 
Furthermore, we observe that the approximation provided by a convex relaxation approach differs significantly from $A^\star_k$ in all the considered settings, as indicated by the error norm for $k< m$. This suggests that the theoretical conditions necessary to recover $A^\star_k$ by convex relaxation do not hold here.
Finally, as expected in the case where $\mathrm{rank}(\XXX)=m$,  the linear operator used to generate the snapshots  is accurately recovered  by our optimal approximation and truncated DMD, TLS DMD and ADMM for $k \ge m$. In the case where $\mathrm{rank}(\XXX)=m-6$, we observe  (in good agreement with Theorem~\ref{prop22})  that the optimal approximation error is lower bounded by $\| \BBB (I_m-\mathbb{P}_{\AAA^{\intercal}})\|_F^2$ for $k \ge m$.

The eigenvalue error plots show that low-rank projected DMD and TLS DMD are optimal respectively in {\it setting~i)} and {\it setting~ii)-~iii)}, as long as $\AAA$ is full rank. However, in other circumstances these methods are sub-optimal for $k<m$. Interestingly, we observe that eigenvalues are well estimated by all the methods in all settings when $k\geq m$,  although low-rank projected DMD exhibits a high reconstruction error in {\it settings~ii)-~iii)}.  These toy experiments show that even if eigenvalues are well estimated, related eigenvectors can be flawed.   \vspace{-0.15cm}

\subsection{Physical Experiments}\label{sec:phys}

Rayleigh-B\'enard model \cite{chandrasekhar2013hydrodynamic} constitutes a benchmark  for convective system in geophysics. Convection is driven by  two coupled  partial differential equations ruling the evolution of temperature and vorticity. After discretization of these equations on the cell $[0,1]\times[0,1/2]$, we obtain  a discrete system of the form of~\eqref{eq:model_init} with $x_t\in \Rr^{n}$, $n=1024$. The regime of the convective system is parametrized by two quantities: the Rayleigh  number $ \operatorname{Ra}\in \Rr_+$ and the Prandtl number $\operatorname{Pr}\in \Rr_+$. The  initial condition $\theta=h(\vartheta)$ is  parametrized by a vector $\vartheta \in \Rr^4$, through the non-linear function $h:\Rr^4 \to \Rr^{1024}$, see details in~\cite{HeasHerzet17}. 
Entries of the vector  $\vartheta$  are  sampled uniformly on (a bounded sub-domain of) $\mathbb{R}^4$, yielding the samples $\{\vartheta_i\}_i$. Then the first eigenvectors of the proper orthogonal decomposition of the set  $\{h(\vartheta_i)\}_i$ are used to form an hyper-cube of  dimension $d=10$. Finally, the set of initial conditions  $\{\theta_i\}_i$ are  obtained by uniform sampling  on this hyper-cube. 
%Entries of the vector in $\Rr^4$  are randomly sampled, implying that the span of the  set of initial conditions is an hyper-cube, whose  dimension is related to the image of function $h$. We set the dimension of this hyper-cube to $d=10$.
For a particular parametrization of the initial condition, $ \operatorname{Ra}$ and $ \operatorname{Pr}$,  the non-linear Rayleigh-B\'enard system can simplify into a linear  
 Taylor vortex evolution~\cite{Taylor37}.   A precise description of the Rayleigh-B\'enard model, its parametrization and discretization is provided in~\cite{HeasHerzet17}.
Three datasets of $m=50$ snapshots of the discretized system trajectories are computed by numerical simulation:  \vspace{-0.15cm}

\begin{itemize}
\item  {\it {setting $iv)$}}: \,\,$N=50$  trajectories of the linear Taylor vortex with $T=2$,
\item  {\it {setting $v)$}}: \,\, $N=5$  trajectories    of the linear Taylor vortex with $T=11$,
\item  {\it {setting $vi)$}}: \,\, $N=5$   trajectories  of the non-linear Rayleigh-B\'enard system with $T=11$. \vspace{-0.25cm}
\end{itemize}

\begin{figure}[h!]
\centering\vspace{-2.5cm}
\begin{tabular}{cc}
\includegraphics[width=0.525\columnwidth]{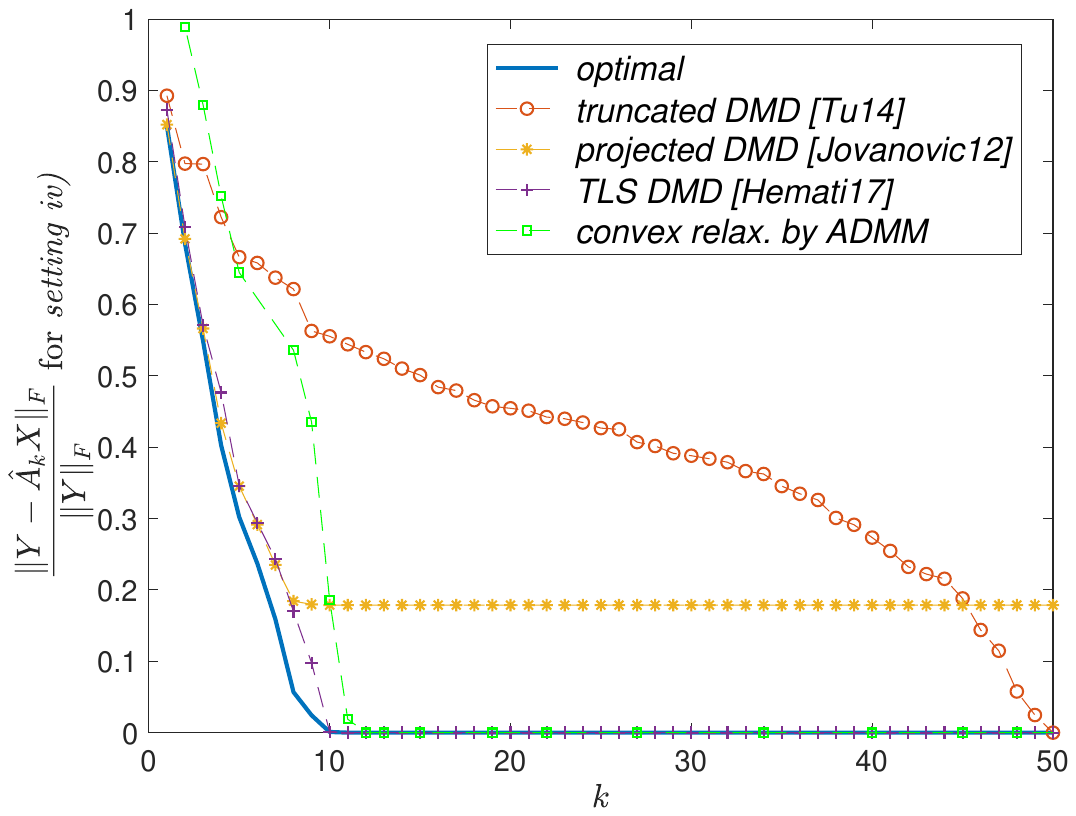}&\hspace{-1.cm}
\includegraphics[width=0.525\columnwidth]{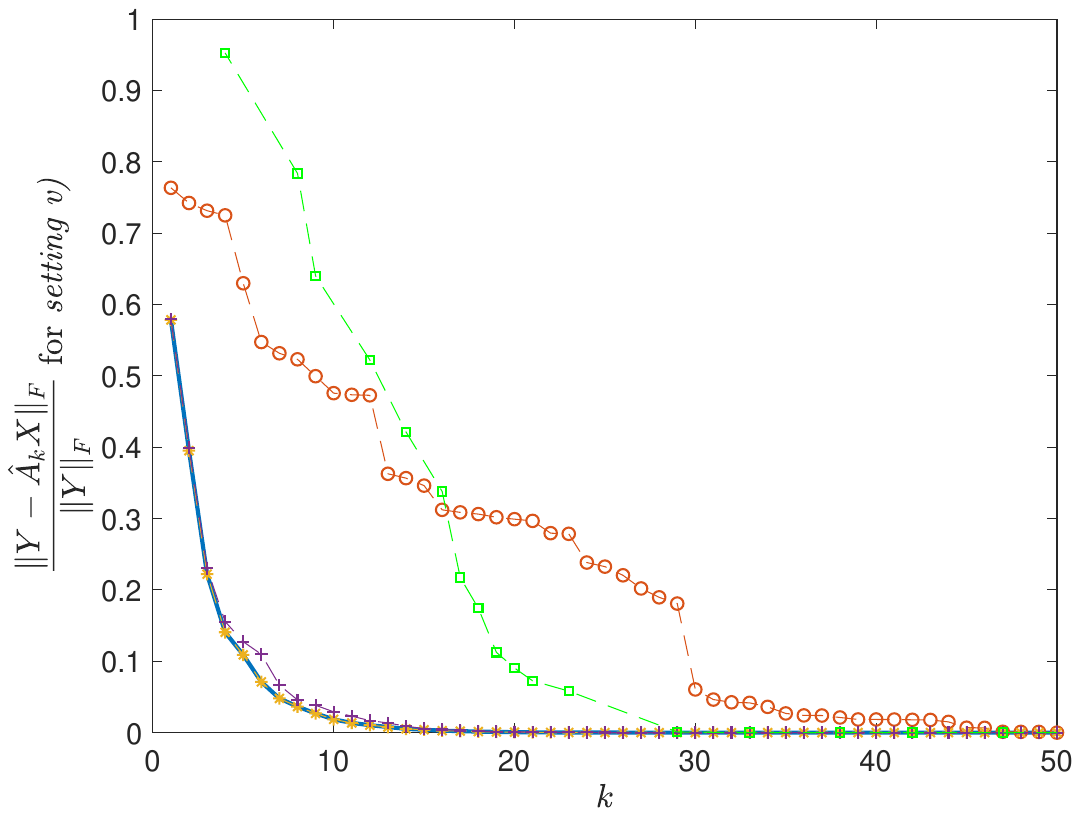}\vspace{-4.5cm}\\
\multicolumn{2}{c}{\includegraphics[width=0.525\columnwidth]{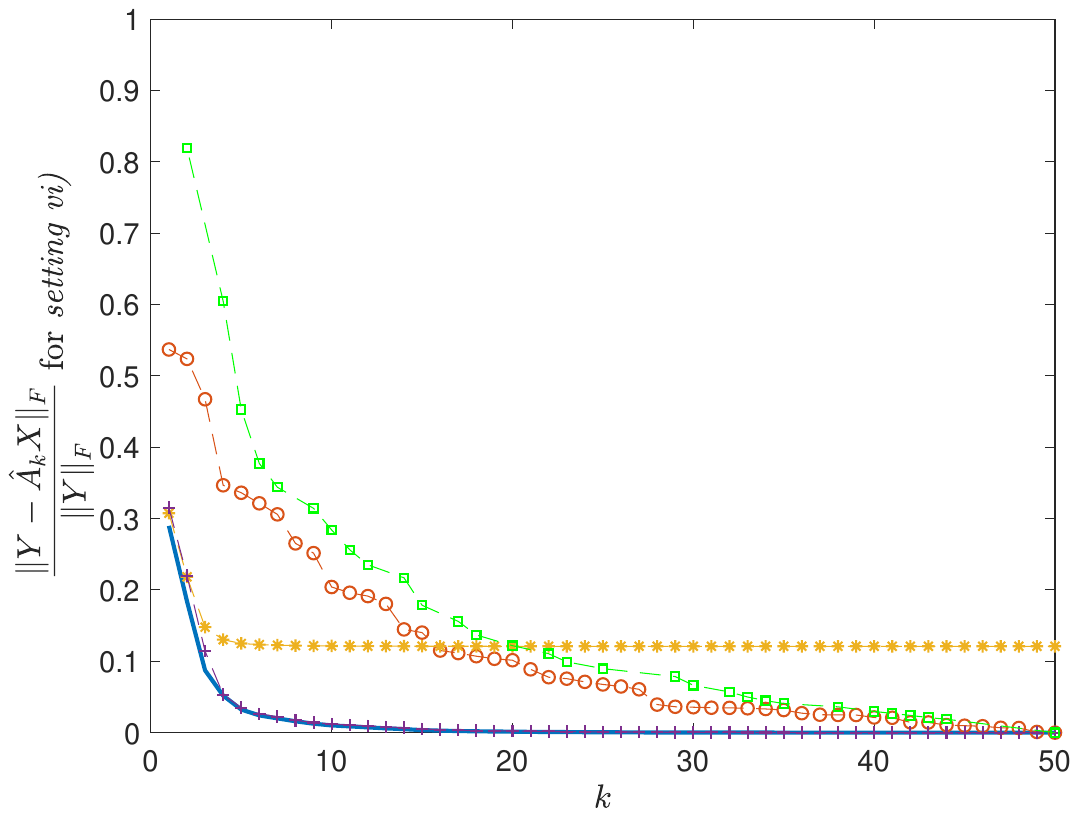}}\vspace{-2.5cm}
\end{tabular}
\caption{\small Reconstruction error norm  as a function of $k$ for   {\it settings} $iv)$, $v)$ and $vi)$ using our optimal approximation or state-of-the-art approximations. See details in Section~\ref{sec:phys}. \vspace{-0.45cm}\label{fig:2}}
\end{figure}

%\subsubsection{Results}
The performance of the different algorithms in terms of reconstruction error is plotted as a function of $k$ in Figure \ref{fig:2}  in the case of these three settings.  
We first comment on results obtained in {\it setting $iv)$}. The error obtained by the optimal approximation in this linear setting with $T=2$ appears to vanish for $k\ge d$, \ie a dimensionality greater than the initial condition dimensionality.  This is consistent with  Theorem~\ref{prop22}. Indeed,   we know by the definition \eqref{eq:defZZZ} that $\textrm{rank}(\ZZZ)\le \textrm{rank}(\BBB)$; then, dealing with a linear model  implies that $ \textrm{rank}(\BBB) \le\textrm{rank}(\AAA)$; as $T=2$  we  have $\textrm{rank}(\AAA)=d$ and therefore $\textrm{rank}(\ZZZ)\le d$ which yields  $\sum_{i=d+1}^m \sigma_{\ZZZ,i}^2=0$; in addition we observe that the term $\| \BBB (I_m-\mathbb{P}_{\AAA^{\intercal}})\|_F^2 \approx 2.e-8$ can  be neglected in our experiments.   It follows from the theorem that the optimal error  vanishes for $k\ge d$.  The approximation by truncated DMD is associated to an important error which vanishes only for $k=m$, \ie for a dimensionality equal to  the number of snapshots. Concerning the approximation by low-rank  projected DMD, it produces a fairly good  solution up to $k \le 8$. However, the approximation becomes clearly sub-optimal  for greater dimensions and produces  an non-negligible error, even for large values of $k$. TLS DMD yields  for $k< 10$ an approximation   slightly less accurate than the one we propose, while the performances of the two methods are indistinguishable for $k\ge10$.  The approach by convex relaxation produces fairly good results, however with a performance significantly lower than other state-of-the-art methods.
 
In {\it setting} $v)$, we have longer sequences  ($T>2$) so that we have   $\textrm{rank}(\AAA)\ge d$.  Although the dynamic is linear, no conclusion can be drawn anymore from  Theorem~\ref{prop22}, except  that the  optimal approximation vanishes for $k\ge m$.   However,  our optimal approximation yields an error which nearly vanishes for $k \ge d$. This shows that, for this linear model,   trajectories concentrate near the subspace spanned by the initial condition. This explains the quasi-optimality of the approximation by low-rank  projected DMD, which relies on a strong assumption of linear dependence of snapshots. An approximation by truncated DMD is again clearly sub-optimal and behaves analogously to {\it setting}~$iv)$.  The performance of TLS DMD is in this setting nearly optimal, while  convex relaxation is disappointing up to some extent.
 
 In the more realistic geophysical {\it setting} $vi)$, we see that  the optimal performance achieved by our approximation is  far from being reached by  approximations by truncated of low-rank  projected DMD. Nevertheless, the approximation obtained by TLS DMD is again nearly optimal. As in the linear settings, we observe that the optimal error is small for $k \ge d $. On the other hand, we clearly notice that the assumption used in the approximation by low-rank  projected DMD does not hold for this non-linear models and produces an important error, even for large values of $k$. We  observe  the poor performance of an approximation by truncated DMD or using convex relaxation.   \vspace{-0.15cm}
 
 %\remPH{Dire que la solution PQ ne change pas avec le modele non stationnaire~\eqref{eq:RBLinear}.}\remPH{A finir .....}
    
  \subsection{Robustness to Noise}\label{sec:robust}

\begin{figure}[h!]
\centering\vspace{-2.5cm}
\begin{tabular}{cc}
\includegraphics[width=0.525\columnwidth]{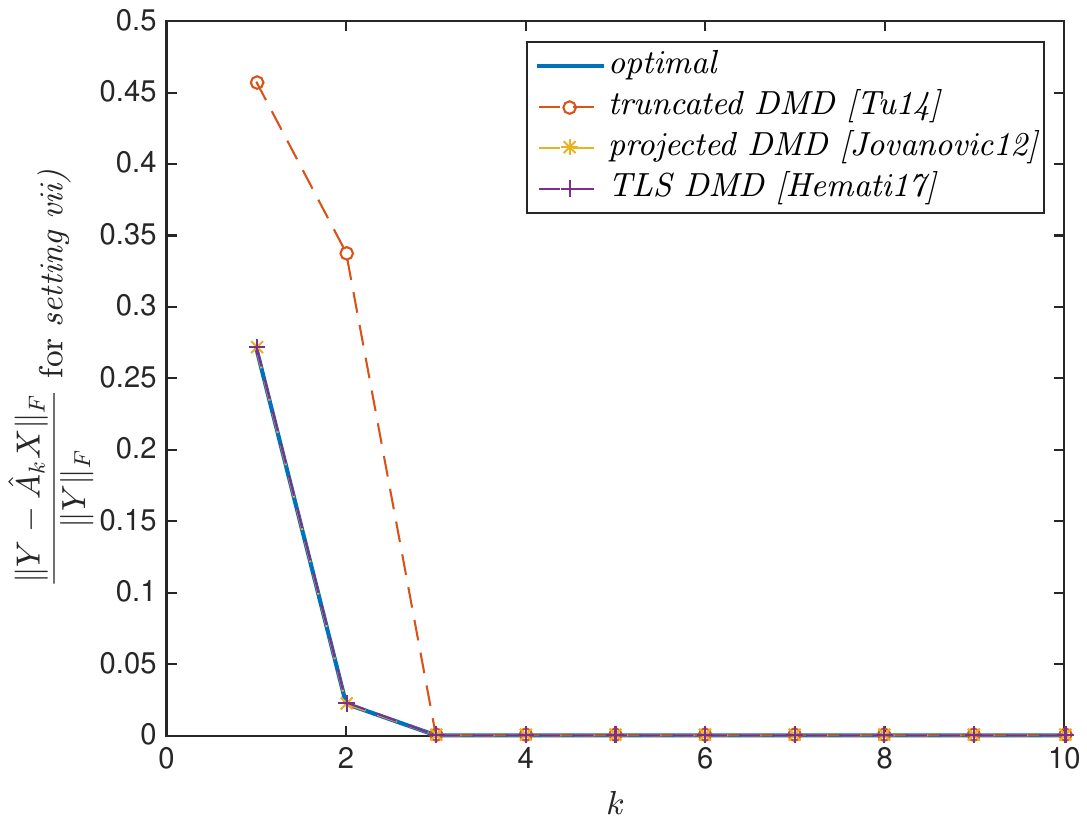}&\hspace{-1.cm}
\includegraphics[width=0.525\columnwidth]{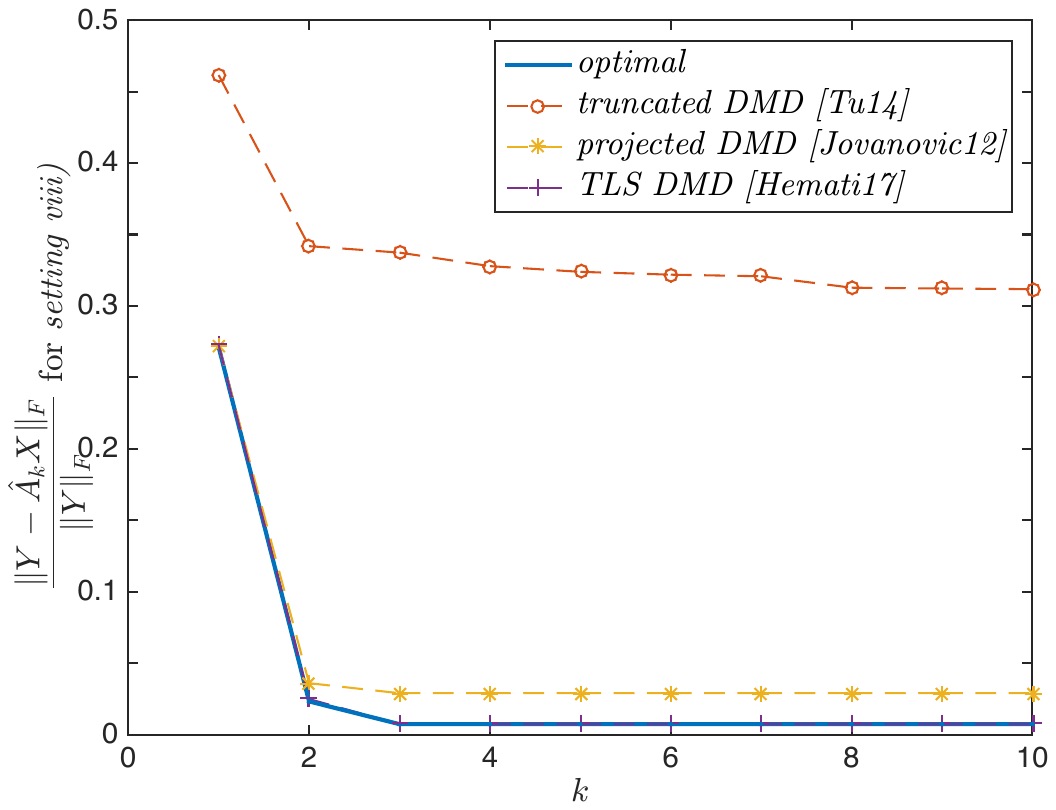}\vspace{-2.cm}\\
\end{tabular}
\caption{\small Error norm    as a function of  $k$ for   {\it setting} $vii)$ and $viii)$  using our optimal approximation or state-of-the-art approximations. See details in Section~\ref{sec:robust}.  %Setting $i)$  ({above}) and $ii)$ (middle)  imply both a linear model but the former satisfies the snapshots linear dependence assumption. Setting $iii)$ ({below}) implements a non-linear model. We evaluate 3  algorithms:  {method $a)$} is the proposed optimal algorithm, {method b)} provides the rank-$k$ SVD approximation of the unconstrained solution given in \cite{Tu2014391} and {{\it method c)}} is the low-rank projected DMD method proposed in \cite{Jovanovic12}. See details in Section~\ref{sec:numEval}. 
\vspace{-0.3cm}\label{fig:3}}
\end{figure}

\begin{figure}[t!]
\centering
\vspace{-0.cm}
\begin{tabular}{cc|cc|cc}
\multicolumn{6}{c}{\footnotesize{Truth}}\\
\hline 
~ \\
\includegraphics[height=0.2\columnwidth]{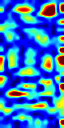}&\includegraphics[height=0.2\columnwidth]{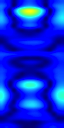}&
\includegraphics[height=0.2\columnwidth]{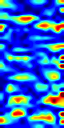}&\includegraphics[height=0.2\columnwidth]{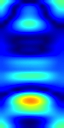}&
\includegraphics[height=0.2\columnwidth]{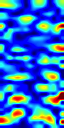}&\includegraphics[height=0.2\columnwidth]{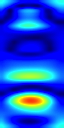}\\
\\
\multicolumn{6}{c}{\footnotesize{Optimal}}\\
\hline 
~ \\
\includegraphics[height=0.2\columnwidth]{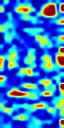}&\includegraphics[height=0.2\columnwidth]{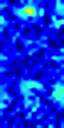}&
\includegraphics[height=0.2\columnwidth]{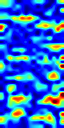}&\includegraphics[height=0.2\columnwidth]{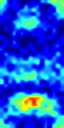}&
\includegraphics[height=0.2\columnwidth]{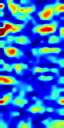}&\includegraphics[height=0.2\columnwidth]{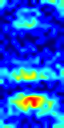}\\
\\
\multicolumn{6}{c}{\footnotesize{Truncated DMD}}\\
\hline 
~ \\
\includegraphics[height=0.2\columnwidth]{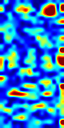}&\includegraphics[height=0.2\columnwidth]{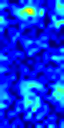}&
\includegraphics[height=0.2\columnwidth]{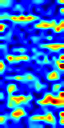}&\includegraphics[height=0.2\columnwidth]{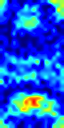}&
\includegraphics[height=0.2\columnwidth]{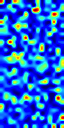}&\includegraphics[height=0.2\columnwidth]{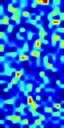}\\
\\
\multicolumn{6}{c}{\footnotesize{low-rank Projected DMD}}\\
\hline 
~ \\
\includegraphics[height=0.2\columnwidth]{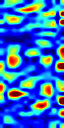}&\includegraphics[height=0.2\columnwidth]{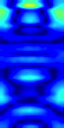}&
\includegraphics[height=0.2\columnwidth]{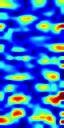}&\includegraphics[height=0.2\columnwidth]{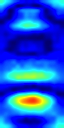}&
\includegraphics[height=0.2\columnwidth]{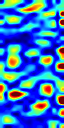}&\includegraphics[height=0.2\columnwidth]{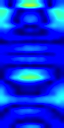}\\
\\
\multicolumn{6}{c}{\footnotesize{TLS DMD}}\\
\hline 
~ \\
\includegraphics[height=0.2\columnwidth]{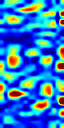}&\includegraphics[height=0.2\columnwidth]{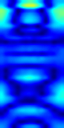}&
\includegraphics[height=0.2\columnwidth]{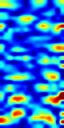}&\includegraphics[height=0.2\columnwidth]{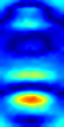}&
\includegraphics[height=0.2\columnwidth]{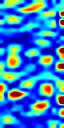}&\includegraphics[height=0.2\columnwidth]{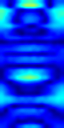}\\
~ \\

\multicolumn{2}{c}{$\zeta_1$}&\multicolumn{2}{c}{$\zeta_2$}&\multicolumn{2}{c}{$\zeta_3$} \vspace{-0.cm}\\

\end{tabular}
\caption{\small Amplitudes related to  temperature (left columns) and vorticity (right columns) of the  right eigenvectors of matrix $F$ of rank $3$ defined in \eqref{GNoiseRobust}: ground truth and  estimation obtained  in the noisy setting \textit{viii)}  with our optimal approximation, with an approximation by truncated DMD, with an approximation by projected DMD  or with an approximation by total-least square DMD. See details in Section~\ref{sec:robust}. 
\vspace{-0.3cm}\label{fig:4}}
\end{figure}

  In the following, we intend to evaluate the ability of the different methods to extract  the eigenvectors in~\eqref{eq:koopman1}    in the presence of noise. To this aim,
we build  a new dataset of $N=5$  long trajectories with $T=11$  (so that we get  $m=50$ snapshots) satisfying~\eqref{eq:koopman1} with $r=3$. The eigenvectors and eigenvalues in the set $\{(\xi_i, \zeta_i, \lambda_i)\}_{ i=1}^3$ are computed using Algorithm~\ref{algo:2}  in the context of {\it setting} $vi)$.  In other words,
matrices $\AAA$ and $\BBB$ are generated using~\eqref{eq:model_init} %from a random initial condition $\theta$ 
and the model $f_t(x_{t-1})=F x_{t-1}$ where 
\begin{align}\label{GNoiseRobust}
F=\begin{pmatrix}\zeta_1&\zeta_2&\zeta_3\end{pmatrix}\textrm{diag}(\lambda_1,\lambda_2,\lambda_3)\begin{pmatrix}\xi_1&\xi_2&\xi_3\end{pmatrix}^\intercal.
\end{align}
We then consider the two following snapshots configurations:

\begin{itemize}
\item  {{{\it setting} $vii)$}}: \,\,  the original matrices $\AAA$ and $\BBB$,
\item  {{{\it setting} $viii)$}}: \,\, a noisy version of matrices $\AAA$ and $\BBB$, where we have corrupted the snapshots with a zero-mean Gaussian noise so that the peak signal-to-noise ratio\footnote{The peak signal-to-noise ratio is defined as $20 \log_{10} \frac{\max_{t,i}\|x_t(\theta_i)\|_\infty}{ \sigma}$, where $\sigma$ denotes the standard deviation of the standard normal distribution.} is $20$ dB.  
\end{itemize}

Results  are displayed in Figures \ref{fig:3} and \ref{fig:4}. As expected the optimal approximation error vanishes in the noiseless setting in the case where $k \ge r$. We  observe only a slight increase   of the error   in the presence of  noise.   This shows  the robustness of the proposed method. Besides, we remark that the error norm obtained with TLS DMD or with the proposed optimal approach are indistinguishable in both, the noiseless and noisy settings. The approximation by low-rank  projected DMD  in the noiseless case reproduces almost exactly the optimal behaviour, while its performance slightly deteriorates for  $k \ge 2$ in the noisy case. The quasi-optimal performance of this method in the noiseless setting  shows that the assumption of linear dependence of snapshots is nearly valid. This assumption no longer holds when snapshots are corrupted by noise.  The approximation by truncated DMD is clearly sub-optimal  in the noiseless setting. More importantly, the performance of this method becomes dramatic in the presence of noise: the error difference with the optimal one is of the order of more than a decade. 

The deterioration is clearly visible  in Figure~\ref{fig:4}. This figure displays the different approximations of eigenvector $\zeta_3$ re-arranged in the form of a spatial map.  In the presence of noise, the spatial structure of $\zeta_3$ is completely rubbed out using an approximation by truncated DMD. Eigenvector $\zeta_3$  is fairly recovered using our optimal approximation and roughly estimated using an approximation by low-rank  projected DMD. Surprisingly, although  TLS DMD  yields a quasi-optimal approximation error norm, the structure of eigenvector $\zeta_3$ is completely rubbed out, in a very similar manner to low-rank  projected DMD. % This shows the importance of solving the constrained problem~\eqref{eq:prob1} rather than truncating the solution of the unconstrained  problem~\eqref{eq:prob1_unconst}.
  
  \vspace{-0.25cm}
  
\section{Conclusion}
%Following recent attempts to characterise  low-rank DMD  approximations, this paper provides a closed-form solution  to this nonconvex optimization problem. To the best of our knowledge,  state-of-the-art methods are all  sub-optimal. The paper further proposes effective algorithms based on SVD to solve this problem and run the reduced model. %, making our contribution significant, with  potentially an important impact in applicative domains.
%Our numerical experiments attest that the proposed algorithm is more accurate than state-of-the-art methods. In particular, we illustrate the fact that  simply truncating the full-rank DMD solution, or  exploiting a strong assumption of linear dependence of snapshots is insufficient.   

This work shows that we can compute in polynomial time an optimal solution of the non-convex problem related to low-rank linear approximation. 
As shown in Theorem~\ref{prop22}, a closed-form  solution is in fact the orthogonal projection of the unconstrained problem solution  onto a specific low-dimensional  subspace. 
%This subspace is the span of the first $k$ left singular vectors of a matrix $\ZZZ$, which is defined as the multiplication of $\BBB$  by the projector onto the span of the rows of $\AAA$.  
The theorem also provides a closed-form  characterization of the optimal approximation error.
% The expression of the  $\ell_2$-norm of the error is in fact closed-form and  depends on the singular values of $\BBB$ and $\ZZZ$ and on the scalar product between right singular vectors of $\AAA$ and $\BBB$.   
Based on these results, we  show in Proposition~\ref{rem:2} that  the EVD of this optimal solution can be obtained directly from the snapshots  with a complexity of $\mathcal{O}(m^2(m+n))$. This off-line complexity is the same as for state-of-the-art sub-optimal methods. 
Finally, we illustrate through numerical simulations in synthetic and physical setups, the gain brought by using this optimal approximation. \vspace{-0.2cm}

\appendix

\section{Proof of Theorem~\ref{prop22}}\label{app:Theorem}

We begin by showing the first part of the theorem, namely that  %assuming $\textrm{rank}(\AAA)\ge k$,
  $A_k^\star={U}_{\ZZZ,k}  {{U}_{\ZZZ,k}}^\intercal  \BBB \AAA^{\dagger}$ is a  solution of 
 \eqref{eq:prob1}.
We first prove in this paragraph  the existence of a minimizer   of \eqref{eq:prob1}. Let us show that we can restrict our attention to a minimization problem over the set $$\mathcal{A}=\{\tilde A \in \Rr^{n \times n}  : \textrm{rank}(\tilde A) \le k, \textrm{Im}(\tilde A^\intercal) \subseteq \textrm{Im}(\AAA)\}.$$  Indeed, any matrix  $ A \in \{\tilde A \in \Rr^{n \times n}  : \textrm{rank}(\tilde A) \le k\}$ can be decomposed in two  components: $ A= A^\parallel+ A^\perp$ where  $ A^\parallel$ belongs to the set  $\mathcal{A}$,  such that  columns of $A^\parallel$ are orthogonal to those of $A^\perp$, \ie $ A^\perp ( A^\parallel)^\intercal=0$. From this construction, we have that rows of $A^\perp$ are orthogonal to rows of  $\AAA$.
 Using this decomposition, we thus have that $\| \BBB -  A \AAA\|_F^2=\| \BBB -  A^\parallel \AAA\|_F^2$. Moreover, because of this orthogonal property, we have that $  \textrm{rank}( A)=\textrm{rank}( A^\parallel) +\textrm{rank}( A^\perp) $  so  that $ \textrm{rank}( A^\parallel) \le \textrm{rank}( A)$. In consequence, if $ A$ is a minimizer  of \eqref{eq:prob1}, then $ A^\parallel$ is also a minimizer since it leads to same value of the cost function and since it is admissible: $\textrm{rank}( A^\parallel) \le \textrm{rank}( A) \le k$.  Therefore, it is sufficient to find a minimizer over the set $\mathcal{A}$.

Now, according to the  Weierstrass' theorem \cite[Proposition A.8]{bertsekas1995nonlinear}, the existence is guaranteed if the admissible set $\mathcal{A}$ is closed and the objective function $\| \BBB - A \AAA\|_F^2$ is coercive. Let us prove these two properties. We first  show that $\mathcal{A}$ is closed. According to \cite[Lemma 2.4]{hackbusch2012tensor}, the set of low-rank matrices is closed. Moreover, it is well-known that  a linear subspace of a normed   finite-dimensional  vector space is closed~\cite[Chapter 7.2]{auliac2005mathematiques}, so that the set of matrices $\mathcal{A}=\{\tilde A \in \Rr^{n \times n}  : \textrm{Im}(\tilde A^\intercal) \subseteq \textrm{Im}(\AAA)\}$   is closed.  Since  $\mathcal{A}$ is the intersection of two closed sets,  we deduce that $\mathcal{A}$ is closed. Next, we show coercivity. Let us consider the SVD of any $A\in \mathcal{A}$: $A=U_A\Sigma_A V_A^\intercal$, where $\Sigma_A=\textrm{diag}(\sigma_{A,1}\cdots\sigma_{A,k})$. % with at most $k$ non-zero singular values denoted by the $\sigma_{A,i}$'s. 
From the definition of the Frobenius norm, we have for any  $A \in \mathcal{A}$,  $ \| A\|_F =( \sum_{i=1}^k\sigma_{A,i}^2)^{1/2}  $.   
%
%the assumptions $\textrm{rank}(\AAA) \neq 0$ and $\textrm{rank}(\BBB) \neq 0$ imply that 
%$$
%\min_{A \in \Rr^{n \times n}  : \textrm{rank}(A) \le k}\| \BBB - A \AAA\|_F^2 < \| \BBB\|_F^2,
%$$
%so that a matrix full of zeros can not be a solution  of \eqref{eq:prob1}. We deduce that matrices with all their lines orthogonal to $\textrm{Im}(\AAA)$ are not solution of \eqref{eq:prob1}. We thus have 
%$$
%\min_{A \in \Rr^{n \times n}  : \textrm{rank}(A) \le k}\| \BBB - A \AAA\|_F^2 =\min_{A \in {K}}\| \BBB - A \AAA\|_F^2.
%$$ 
We have  that $\| A\|_F  \to \infty$ if a non-empty subset of singular values, say  $\{\sigma_{A,j}\}_{j \in \mathcal{J}}$, tend to infinity. Therefore, we  have 
\begin{align*}
\lim_{ \| A\|_F \to \infty :A \in \mathcal{A} } \| \BBB -  A \AAA\|_F^2 
&=  \lim_{\| A\|_F \to \infty: A \in \mathcal{A}  } \| \BBB\|^2_F -2 \,\textrm{trace}(Y^\intercal A \AAA )+ \| A \AAA\|_F^2,  \\
&=  \lim_{ \| A\|_F \to \infty :A \in \mathcal{A} } \| A \AAA\|_F^2
=  \lim_{\| A\|_F \to \infty : A \in \mathcal{A} } \| \Sigma_A V_A^\intercal \AAA\|_F^2,  \\
&=    \lim_{\sigma_{A,j} \to \infty : A \in \mathcal{A},j \in \mathcal{J}} \sum_{j=1}^n \sigma_{A,j}^2  \|   \AAA^\intercal v_A^{j} \|_2^2 
     =  \infty.
    \end{align*}
The second equality is obtained because the dominant term when $\| A\|_F \to \infty$  is  the quadratic one $ \| A \AAA\|_F^2$.  The third equality follows from the invariance of the Frobenius norm to unitary transforms while the last equality is obtained noticing that $ \|   \AAA^\intercal v_A^{j} \|_2 \neq 0$ because $ v_A^{j} \in  \textrm{Im}(\AAA)$ since $A \in \mathcal{A}$. 
%Obviously we also obtain this limit in the case wherethere is more than one singular value tending to infinity. 
This shows that the objective function is coercive over the closed set $\mathcal{A}$. Thus, using the Weierstrass' theorem, this shows   the existence of a minimizer of \eqref{eq:prob1} in $\mathcal{A}$ and thus in $\{\tilde A \in \Rr^{n \times n}  : \textrm{rank}(\tilde A) \le k\}$. 
%Therefore, we deduce from the above discussion that there  exists a minimizer  of \eqref{eq:prob1} satisfying the  first-order optimality condition\footnote{
%Although we have shown that the minimizer of the objective over the set $\mathcal{A}$ are also solutions of problem \eqref{eq:prob1}, 
We will no longer restrict our attention to the domain $\mathcal{A}$ in the following and come back to the original problem \eqref{eq:prob1} implying the  set of low-rank matrices.
%
%}.

Next, problem~\eqref{eq:prob1} can be rewritten as the unconstrained minimization  
\begin{align}\label{eq:prob1unconst}
		A_k^\star \in &\argmin_{A=PQ^\intercal: P,Q \in \Rr^{n \times k}} \|\BBB -A \AAA \|^2_F.
\end{align}
In the following we will use the  first-order optimality  condition of problem \eqref{eq:prob1unconst} to characterize its minimizers. A closed-form expression for a minimizer will then be obtained be introducing an additional orthonormal property. The first-order optimality condition and the additional orthonormal property are presented in the following lemma, which is proven in Appendix~\ref{sec:app2}.

  \begin{lemma}\label{rem:py=qx}
 %Assuming $\textrm{rank}(\AAA) \ge k $, there exists a
 Problem \eqref{eq:prob1unconst} admits a solution such that  %satisfying orthogonality contraints \eqref{eq:orthCOnst} 
\begin{align}
&\R^\intercal \R=I_k\label{eq:CNsuppNorm}\\
 &\AAA\BBB^\intercal \R = \AAA \AAA^\intercal \Q\label{eq:CNsupp}. %\\
 %&(\BBB-\R\Q^\intercal\AAA)\AAA^\intercal\Q=0. \label{eq:CNsupp2}
\end{align}
\end{lemma}

To find a  closed-form expression of a minimizer of \eqref{eq:prob1unconst}, we need to rewrite condition~\eqref{eq:CNsupp}. We  prove that this condition is equivalent to 
\begin{align}\label{eq:ii}
\mathbb{P}_{\AAA^\intercal}\BBB^\intercal P=\AAA^\intercal \Q.
\end{align}
Indeed, we  show  by contradiction that \eqref{eq:CNsupp}  implies that,  for any solution of the form $\R\Q^\intercal $, there exists   $Z\in \Rr^{m \times k}$   such that 
%$
%\textrm{im}(\BBB^\intercal P) \subseteq \textrm{im}(\AAA^\intercal )
%$ that is
%\begin{align}\label{eq:ii}
%\AAA^\intercal (\AAA^\intercal )^\dag \BBB^\intercal P=\BBB^\intercal P.
%\end{align}
\begin{align}\label{eq:eqsurZ}
 \mathbb{P}_{\AAA^\intercal}\BBB^\intercal P +Z=\AAA^\intercal \Q,
\end{align}
with  %$ (I_m-\AAA^\dagger \AAA) Z=Z$ (\ie with
 columns of $Z$ in $\ker(\AAA)$.  Indeed, if $ \mathbb{P}_{\AAA^\intercal}\BBB^\intercal P +Z \neq \AAA^\intercal \Q$, then by multiplying both sides on the left by $\AAA$ we obtain $ \mathbb{P}_{\AAA}\AAA\BBB^\intercal P +\AAA Z= \mathbb{P}_{\AAA} \AAA\BBB^\intercal P  \neq\AAA \AAA^\intercal \Q$. Since $ \mathbb{P}_{\AAA}$ is the orthogonal projector onto the subspace spanned by the columns of $\AAA$, the latter relation implies that $ \AAA\BBB^\intercal P \neq\AAA \AAA^\intercal \Q$ which contradicts \eqref{eq:CNsupp}. This proves that \eqref{eq:CNsupp} implies \eqref{eq:eqsurZ}.

Now, since  columns of the two terms in the left-hand side of \eqref{eq:eqsurZ} are orthogonal and since columns of the matrix in the right-hand side are in the image of $\AAA^\intercal$, we deduce that the only admissible choice is $Z$ with  columns  belonging both to $\ker(\AAA)$ and $\textrm{Im}(\AAA^\intercal)$, \ie $Z$ is a matrix full of zeros. Therefore, we obtain the necessary condition \eqref{eq:ii}.

We have shown on the one hand that \eqref{eq:CNsupp} implies \eqref{eq:ii}. On the other hand,  by multiplying on the left both sides of \eqref{eq:ii}  by $\AAA$,  we obtain \eqref{eq:CNsupp} ($\AAA\mathbb{P}_{\AAA^\intercal}=\AAA$ because $\AAA \AAA^\dag$ is the orthogonal projector onto the space spanned by the columns of $\AAA$).  Therefore  the necessary conditions \eqref{eq:CNsupp} and~\eqref{eq:ii} are equivalent.

We are now ready to characterize a minimizer of \eqref{eq:prob1}. According to Lemma \ref{rem:py=qx}, we  have
\begin{align}
\min_{A \in \Rr^{n \times n} : \textrm{rank}(A) \le k }&\| \BBB - A\AAA\|_F^2 \nonumber  \\
=&\min_{( \tilde \R, \tilde\Q) \in \Rr^{n \times k} \times  \Rr^{n \times k} }\| \BBB - \tilde \R\tilde \Q^\intercal \AAA\|_F^2 \quad s.t. \quad 
 \left\{\begin{aligned}
 &\tilde \R^\intercal \tilde \R=I_k\\
 &\AAA\BBB^\intercal\tilde \R = \AAA \AAA^\intercal \tilde\Q\\ %\AAA \BBB^\intercal \tilde \R= \AAA \AAA^\intercal \tilde \Q\\
\end{aligned}\right. ,\label{eq:P0prim0} \\
=&\min_{( \tilde \R, \tilde\Q) \in \Rr^{n \times k} \times  \Rr^{n \times k} }\| \BBB - \tilde \R\tilde \Q^\intercal \AAA\|_F^2 \quad s.t. \quad 
 \left\{\begin{aligned}
 &\tilde \R^\intercal \tilde \R=I_k\\
 & \mathbb{P}_{\AAA^\intercal}\BBB^\intercal \tilde P=\AAA^\intercal \tilde \Q\\ %\AAA \BBB^\intercal \tilde \R= \AAA \AAA^\intercal \tilde \Q\\
\end{aligned}\right. ,\nonumber\\
%=&\min_{ \tilde \R \in \Rr^{n \times k}  }\| \BBB - \tilde \R \tilde\R^\intercal \BBB\|_F^2 \quad s.t. \quad 
%\left\{\begin{aligned}
% &\tilde \R^\intercal \tilde \R=I_k\\
% &\AAA^\intercal (\AAA^\intercal )^\dag \BBB^\intercal \tilde P=\BBB^\intercal \tilde P\\
%\end{aligned}\right.,\nonumber\\
=&\min_{ \tilde \R \in \Rr^{n \times k} }\| \BBB - \tilde \R \tilde\R^\intercal \BBB \mathbb{P}_{\AAA^\intercal}  \|_F^2 \quad s.t. \quad 
 \tilde \R^\intercal \tilde \R=I_k,\label{eq:objFuncOneTerm}\\
=&\min_{ \tilde \R \in \Rr^{n \times k} }\| (\BBB- \tilde \R \tilde\R^\intercal \BBB )\mathbb{P}_{\AAA^\intercal} + \BBB (I_m-\mathbb{P}_{\AAA^\intercal}) \|_F^2 \quad s.t. \quad 
 \tilde \R^\intercal \tilde \R=I_k, \nonumber \\
=&\min_{ \tilde \R \in \Rr^{n \times k} }\| \ZZZ - \tilde \R \tilde\R^\intercal \ZZZ\|_F^2 + \|  \BBB  (I_m-\mathbb{P}_{\AAA^\intercal}) \|_F^2 \quad s.t. \quad 
 \tilde \R^\intercal \tilde \R=I_k. \label{eq:P0prim}
 \end{align}
 The second equality is obtained from the equivalence between~\eqref{eq:CNsupp} and \eqref{eq:ii}. The third equality    is obtained by  introducing the second constraint in the cost function and noticing that projection operators are always symmetric, \ie
%\begin{align*}
$(\mathbb{P}_{\AAA^\intercal})^\intercal= %\AAA^\intercal (\AAA^\intercal)^\dagger=  V_{\AAA}  \Sigma_{\AAA} \Sigma_{\AAA}^\dagger   V_{\AAA}^\intercal = \AAA^\dagger\AAA=
\mathbb{P}_{\AAA^\intercal}, %\label{eq:remSurX}
 %\BBB \AAA^\dagger\AAA ( \AAA^\dagger\AAA )^\intercal  \BBB^\intercal = \BBB \AAA^\dagger\AAA  \BBB^\intercal. \label{eq:remSurXbis}
 %\end{align*}
 $
 while the last equality follows from the definition of $\ZZZ$ given in \eqref{eq:defZZZ} and the orthogonality of the columns of the two terms. 
Problem \eqref{eq:P0prim} is  a proper orthogonal decomposition  problem with the snapshot matrix $\ZZZ $. 
%More explicitly, noticing that $\BBB \AAA^\dagger\AAA ( \BBB\AAA^\dagger\AAA )^\intercal   = \BBB V_{\AAA}  \Sigma_{\AAA} \Sigma_{\AAA}^\dagger   V_{\AAA}^\intercal  \BBB^\intercal =\BBB \AAA^\dagger\AAA  \BBB^\intercal$, 
The solution of this  proper orthogonal decomposition problem  is the matrix ${U}_{\ZZZ,k}$ (with orthonormal columns) defined in Section~\ref{sec:closedSol}, see \eg~\cite[Proposition 6.1]{quarteroni2015reduced}. We thus obtain from \eqref{eq:objFuncOneTerm}  that 
\begin{align}\label{eq:mimimum}
\min_{A \in \Rr^{n \times n} : \textrm{rank}(A) \le k }\| \BBB - A\AAA\|_F^2 =\| \BBB -{U}_{\ZZZ,k}{{U}_{\ZZZ,k}}^\intercal\BBB \mathbb{P}_{\AAA^\intercal} \|_F^2 =\| \BBB - \mathbb{P}_{\ZZZ,k} \BBB \mathbb{P}_{\AAA^\intercal} \|_F^2 . 
\end{align}
 % By the way, we verify  that    ${U}_{\ZZZ,k}$  is  admissible since  eigenvectors of a symmetric matrix are real orthonormal vectors.  
Furthermore, we verify that $A_k^\star={U}_{\ZZZ,k}{W}^\intercal $ with ${W}=(\AAA^\intercal )^\dag\BBB^\intercal {U}_{\ZZZ,k}$ is  a minimizer of~\eqref{eq:prob1unconst}. Indeed, %using property \eqref{eq:ii}, 
 since
$\AAA \AAA^\intercal {W}=\AAA \AAA^\intercal (\AAA^\intercal )^\dag\BBB^\intercal {U}_{\ZZZ,k}= \AAA \BBB^\intercal {U}_{\ZZZ,k}$, we check that $({U}_{\ZZZ,k},{W})$ is admissible for problem \eqref{eq:P0prim0}.
%so as ${U}_{\ZZZ,k}$  since the set $\{(U_{\BBB})_\ell\}_{\ell=1}^k$ is constituted of orthonormal vectors.
We also check using \eqref{eq:ii} that 
$ \| \BBB - {U}_{\ZZZ,k}{W}^\intercal \AAA\|_F^2=\| \BBB - \mathbb{P}_{\ZZZ,k} \BBB\mathbb{P}_{\AAA^\intercal}\|_F^2, $
\ie that $({U}_{\ZZZ,k},{W})$ reaches the minimum given in \eqref{eq:mimimum}. %is a minimizer of \eqref{eq:P0prim0}.  
In consequence, we have shown that  problem \eqref{eq:prob1unconst}, and equivalently problem \eqref{eq:prob1}, admit the minimizer  $A_k^\star={U}_{\ZZZ,k}{W}^\intercal = \mathbb{P}_{\ZZZ,k} \BBB \AAA^{\dagger}$. \\% where columns of matrix ${U}_{\ZZZ,k}$ are the $k$ first eigenvectors of $\BBB\BBB^\intercal $. 

It remains to prove the second part of the theorem, namely the characterization of the approximation error.   The sought result follows from  standard  proper orthogonal decomposition analysis. Indeed, according to  \cite[Proposition 6.1]{quarteroni2015reduced}  the first term of the cost function in \eqref{eq:P0prim} evaluated at $A_k^\star$ is  
$
 \| \ZZZ-  \mathbb{P}_{\ZZZ,k} \ZZZ\|_F^2= \sum_{i=k+1}^m \sigma_{\ZZZ,i}^2.
$
%We    rewrite the second term  of the cost function~\eqref{eq:P0prim} as
%\begin{align}\label{eq:ErrorSecondTerm}
% \|  \BBB  (I_m-\mathbb{P}_{\AAA^\intercal}) \|_F^2&=  \|  \Sigma_{\BBB} V_{\BBB}^\intercal V_{\AAA} (I_m-\Sigma_{\AAA} \Sigma_{\AAA}^\dagger ) V_{\AAA}^\intercal \|_F^2,\nonumber \\
%&=  \|  \Sigma_{\BBB} V_{\BBB}^\intercal V_{\AAA} (I_m-\Sigma_{\AAA} \Sigma_{\AAA}^\dagger ) \|_F^2 = \left\lVert \begin{pmatrix}  \sigma_{\BBB,1} (v_{\BBB}^1)^\intercal \\ \vdots \\ \sigma_{\BBB,m} (v_{\BBB}^m)^\intercal  \end{pmatrix}  \begin{pmatrix} v_{\AAA}^{i^*} & \cdots &  v_{\AAA}^{m}\end{pmatrix}     \right\lVert_F^2, \nonumber\\
%&=  \sum_{i=i^*}^m \left\lVert  \begin{pmatrix}  \sigma_{\BBB,1} (v_{\BBB}^1)^\intercal \\ \vdots \\ \sigma_{\BBB,m} (v_{\BBB}^m)^\intercal  \end{pmatrix}   v_{\AAA}^{i}     \right\lVert_2^2
%= \sum_{i=i^*}^m \sum_{j=1}^{m} \sigma^2_{\BBB,j} ((v_{\BBB}^j)^\intercal  v_{\AAA}^i  )^2.
%\end{align}
%where the  first and second equalities follow from the invariance of the Frobenius norm to unitary transforms, and more precisely  to the  multiplication on the left by $U_{\BBB}^\intercal$ and on the right by  $V_{\AAA}$.  Gathering error contributions \eqref{eq:ErrorFirstTerm} and \eqref{eq:ErrorSecondTerm}, we obtain the sought result.
%$\square$
%% +\sum_{i=\textrm{rank}(\AAA)+1}^m \left( \sum_{j=1}^m \sigma_{\BBB}^j  (v_{\AAA}^i)^\intercal v_{\BBB}^j  \right)^2.
\vspace{-0.25cm}

\section{Proof of Lemma \ref{rem:py=qx}}\label{sec:app2}

We begin by proving that any minimizer   of \eqref{eq:prob1unconst} can be rewritten  as $\R\Q^\intercal $ where $ \R^\intercal \R=I_k$. Indeed, 
%\begin{lemma}
%Any solution of \eqref{eq:prob1} can be rewritten in terms of $(\tilde \R,\tilde \Q)$, where matrice $\tilde \R$  possess  orthonormal columns.\\
%\end{lemma}
the existence of the SVD of $ \tilde A$ for any minimizer $\tilde A \in \Rr^{n \times n }$  guarantees that $$ \| \BBB -   \tilde A \AAA\|^2_F =  \| \BBB -  U_{ \tilde A }\Sigma_{ \tilde A }V^\intercal _{ \tilde A }\AAA\|^2_F,$$
where $U_{ \tilde A } \in \Rr^{n \times k}$   possesses  orthonormal columns. 
Making the identification $ \R=U_{ \tilde A }$ and $ \Q=V_{ \tilde A }\Sigma_{ \tilde A }$ we verify that  $ \| \BBB -   \tilde A \AAA\|^2_F= \| \BBB -   \R  \Q^\intercal \AAA\|^2_F$ and that  $ \R$  possesses  orthonormal columns.
Next,  any solution $PQ^\intercal$ of \eqref{eq:prob1unconst}  should satisfy the first-order optimality  condition with respect to the $j$-th column denoted $q_j$  of matrix $Q$, that is 
\begin{align*}
2[-\AAA \BBB^\intercal p_j + \sum_{i=1}^k (p_i^\intercal p_j) \AAA \AAA^\intercal q_i]=0,
\end{align*}
where  the $j$-th column  of matrix $P$ is denoted $p_j$.
In particular, a solution with $ \R$  possessing  orthonormal columns should satisfy
$
\AAA \BBB^\intercal p_j= \AAA \AAA^\intercal q_j ,
$
or in matrix form
$\AAA \BBB^\intercal \R=\AAA \AAA^\intercal \Q . \quad \square$ 
\vspace{-0.5cm}
\section{Proof of Proposition~\ref{rem:2}}\label{app:prop}

We have $A_k^\star= \mathbb{P}_{\ZZZ,k}  \BBB \AAA^\dagger={U}_{\ZZZ,k} {W}^\intercal$ which implies that
$${W}^\intercal {U}_{\ZZZ,k}={{U}_{\ZZZ,k}}^\intercal  \BBB \AAA^\dagger {U}_{\ZZZ,k}= {{U}_{\ZZZ,k}}^\intercal  \mathbb{P}_{\ZZZ,k}  \BBB \AAA^\dagger {U}_{\ZZZ,k}= {{U}_{\ZZZ,k}}^\intercal {U}_{\ZZZ,k} {W}^\intercal {U}_{\ZZZ,k}.$$
Using the definition of $\zeta_i$'s and $\xi_i$'s in \eqref{eq:defEigenvectors}, since the $w^r_i$'s and $w^\ell_i$'s are the right and left eigenvectors of ${W}^\intercal {U}_{\ZZZ,k}$, we 
%have that  $$\zeta_i=U_{{W}} w^r_i=\frac{1}{\lambda_i}U_{\hat\Q} U_{\hat\Q}^\intercal{U}_{\ZZZ,k} V_{\hat\Q}\Sigma_{\hat\Q} w^r_i,$$ and  $$\xi_i={U}_{\ZZZ,k} w^\ell_i=\frac{1}{\lambda_i} {U}_{\ZZZ,k}{{U}_{\ZZZ,k}}^\intercal{W} w^\ell_i.$$ Therefore, 
 verify that  
$$ A^\star_k \zeta_i= {U}_{\ZZZ,k} {W}^\intercal {U}_{\ZZZ,k} w^r_i= %\frac{1}{\lambda_i}\hat S {W}^\intercal {U}_{\ZZZ,k} w^r_i=
 {U}_{\ZZZ,k} \lambda_i w^r_i =\lambda_i \zeta_i, $$
and that 
$$ (A^\star_k)^\intercal \xi_i=\hat\Q {{U}_{\ZZZ,k}}^\intercal {W} w^\ell_i = {W} \lambda_i w^\ell_i=\lambda_i \xi_i. $$
Finally, $\xi_i^\intercal \zeta_i =1$ is a sufficient condition so that $\xi_i^\intercal A_k^\star \zeta_i =\lambda_i. \quad \square$
 
 \vspace{-0.15cm}

\begin{acknowledgements}
The authors  thank the ``Agence Nationale de la Recherche" (ANR) which partially funded this research through the GERONIMO project (ANR-13-JS03-0002).
\end{acknowledgements}
 \vspace{-0.5cm}
\bibliographystyle{spmpsci}      % mathematics and physical sciences
%\bibliographystyle{spphys}       % APS-like style for physics
%\bibliography{}   % name your BibTeX data base
%\bibliography{./bibtex}
\bibliography{bibtex}
\end{document}